\documentclass{article}

\usepackage{microtype}
\usepackage{graphicx}
\usepackage{subfigure}
\usepackage{booktabs} 

\usepackage{hyperref}
\usepackage{amsmath}
\usepackage{amssymb}
\usepackage{mathtools}
\usepackage{amsthm}

\usepackage[capitalize,noabbrev]{cleveref}
\usepackage[utf8]{inputenc} 
\usepackage[T1]{fontenc}    
\usepackage{url} 
\usepackage{bm}
\usepackage{booktabs}       
\usepackage{amsfonts}       
\usepackage{nicefrac}       
\usepackage{microtype}      
\usepackage{xcolor}    
\usepackage{graphicx}
\usepackage{subfigure} 
\usepackage{amsmath}
\usepackage{amssymb}
\usepackage{makeidx}
\usepackage{multicol}
\usepackage{balance}
\usepackage{enumitem}
\usepackage{array}
\usepackage{comment}
\usepackage{caption}
\usepackage{adjustbox}
\usepackage{amsthm}
\usepackage{mwe}

\usepackage[accepted]{icml2024}

\usepackage[capitalize,noabbrev]{cleveref}

\theoremstyle{plain}
\newtheorem{theorem}{Theorem}[section]
\newtheorem{proposition}[theorem]{Proposition}

\theoremstyle{definition}

\theoremstyle{remark}

\usepackage[textsize=tiny]{todonotes}

\icmltitlerunning{Discovering Symmetry Breaking in 3D Physical Systems with Relaxed Group Convolution}

\begin{document}

\twocolumn[
\icmltitle{Discovering Symmetry Breaking in Physical Systems with \\ Relaxed Group Convolution}




\begin{icmlauthorlist}
\icmlauthor{Rui Wang}{mit}
\icmlauthor{Elyssa Hofgard}{mit}
\icmlauthor{Han Gao}{harvard}
\icmlauthor{Robin Walters}{neu}
\icmlauthor{Tess E. Smidt}{mit}
\end{icmlauthorlist}

\icmlaffiliation{mit}{Massachusetts Institute of Technology}
\icmlaffiliation{harvard}{Harvard University}
\icmlaffiliation{neu}{Northeastern University}

\icmlcorrespondingauthor{Rui Wang}{rayruw@mit.edu}

\icmlkeywords{Machine Learning, ICML}

\vskip 0.3in
]



\printAffiliationsAndNotice{} 

\begin{abstract}
Modeling symmetry breaking is essential for understanding the fundamental changes in the behaviors and properties of physical systems, from microscopic particle interactions to macroscopic phenomena like fluid dynamics and cosmic structures. Thus, identifying sources of asymmetry is an important tool for understanding physical systems. In this paper, we focus on learning asymmetries of data using relaxed group convolutions.  We provide both theoretical and empirical evidence that this flexible convolution technique allows the model to maintain the highest level of equivariance that is consistent with data and discover the subtle symmetry-breaking factors in various physical systems. We employ various relaxed group convolution architectures to uncover various symmetry-breaking factors that are interpretable and physically meaningful in different physical systems, including the phase transition of crystal structure, the isotropy and homogeneity breaking in turbulent flow, and the time-reversal symmetry breaking in pendulum systems. 
\end{abstract}

\section{Introduction}
Symmetry play pivotal roles in the advancement of deep learning \citep{zhang1988shift, kondor2018generalization, bronstein2021geometric, Cohen2016Group, weiler2019e2cnn, zaheer2017deep, cohen2016steerable}. Specifically, equivariant convolution \citep{cohen2016steerable, esteves2018learning} and graph neural networks \citep{Garcia2021EN, thomas2018tensor, cohen2019gauge}, which integrate symmetries into the design of the architectures, have demonstrated notable success in modeling complex data. By constructing a model inherently equivariant to the symmetry group governing transformations of a physical system, we ensure automatic generalization across these transformations. This enhances not only the model's robustness to distributional shifts but also its sample efficiency \citep{geiger2022e3nn, Walters2021ECCO, wang2020incorporating, helwig2023group, liao2023equiformerv2}. 

The symmetry of the outputs of any equivariant model will have equal or higher symmetry than the inputs to the model \cite{smidt2021finding}. This suggests that equivariant models might be too restrictive to be useful in describing symmetry breaking. However, understanding the effects that give rise to symmetry breaking is key for predicting the behavior of complex systems, from the intricacies of particle interactions to the dynamics of fluid flows \cite{fahle2001symmetry, weinberg1976implications, beekman2019introduction}. Symmetry breaking occurs in two forms: explicit, when the governing equations of a physical system lack symmetry, and spontaneous, when a symmetric system transitions to a lower-energy state with lower symmetry \cite{strocchi2005symmetry}. Depending how this phenomenon manifests in the data, it can be categorized as either functional or distributional. Hence, our objective is to construct models capable of identifying asymmetries across these scenarios.


While traditional equivariant models may be unable to describe some forms of symmetry breaking, \citet{smidt2021finding} shows that additional trainable equivariant symmetry-breaking inputs can be used to learn asymmetries throughout training with an equivariant model. Relaxed group convolution \cite{wang2022approximately} furthers this idea by incorporating trainable relaxed weights to ease the strict weight-sharing scheme in equivariant convolutions. In \citet{wang2022approximately}, the authors empirically show the ability of relaxed group convolutions to model approximate symmetries, but they do not provide any theoretical guarantees or insights. Our work advances the understanding of relaxed group convolutions by both theoretically and empirically showing that they can consistently capture the correct symmetry inductive biases from the data in all scenarios of symmetry breaking. We also demonstrate that after training, the relaxed weights can be used to discover both global and local symmetry-breaking factors in various physical systems. Depending on the type of asymmetry that we aim to learn, the relaxed weights can be analyzed and interpreted in a variety of ways.


Our key contributions are summarized as follows:
\vspace{-7pt}
\begin{itemize}[leftmargin=15pt, itemsep=0.5pt]
    \item Theoretical and empirical evidence of the ability of relaxed weights to detect symmetry breaking and quantify the degree of this breaking. 
    \item Categorization of different types of symmetry breaking that may be seen in physical data and how relaxed group convolutions can be used in different cases.
    \item Discovery of symmetry-breaking factors in diverse physical systems using relaxed group convolution, including phase transitions of crystal structure, isotropy and homogeneity breaking in turbulence, and time-reversal symmetry breaking in pendulum systems. 
    \item Superior performance of relaxed group convolution compared to baselines with no symmetry and with strict symmetry biases on the task of fluid super-resolution for 3D channel flow and isotropic flow. 
\end{itemize}



\section{Background}
\subsection{Equivariant Neural Networks}\label{def:equivariance}

We say a function $f \colon X \to Y$  is \textbf{$G$-equivariant} if 
\begin{equation}
    f( \rho_{\text{in}}(g)(x)) = \rho_{\text{out}}(g) f(x) \label{eq:strictequivariance}
\end{equation}
for all $x \in X$ and $g \in G$, where $\rho_{\text{in}}$ is the input representation of $G$ acting on the vector space $X$ and $\rho_{\text{out}}$ is the output representation  of $G$ on the vector space $Y$. 
Equivariant deep learning often involves techniques like weight sharing and weight tying across group elements. Our focus is on group convolution \citep{Cohen2016Group} and relaxed group convolution \citep{wang2022approximately} and their application to data defined over a regular square or cubic grid.

We formally distinguish between invariant (scalar) weights used in group convolutions and equivariant (non-scalar) weights used in relaxed group convolutions. Let $\mathbb{R}^p$ be the space of model weights. The group $G$ not only acts on input and output spaces, it may also act on weight space $\mathbb{R}^p$ by $\rho_w: G \to \mathrm{GL}_p(\mathbb{R})$. The weight space then decomposes as a direct sum of different representations of $G$.
We say that the weights are invariant if they transform as the trivial representation and equivariant if they transform as a direct sum of non-trivial representations of $G$ (e.g. the regular representation) \cite{Dresselhaus2008}. See Appendix \ref{app:background} for more background on equivariant neural networks, group theory, and representation theory.

\subsection{Symmetry Breaking}
\begin{figure*}[htb!]
	\centering
	\includegraphics[width=0.8\textwidth]{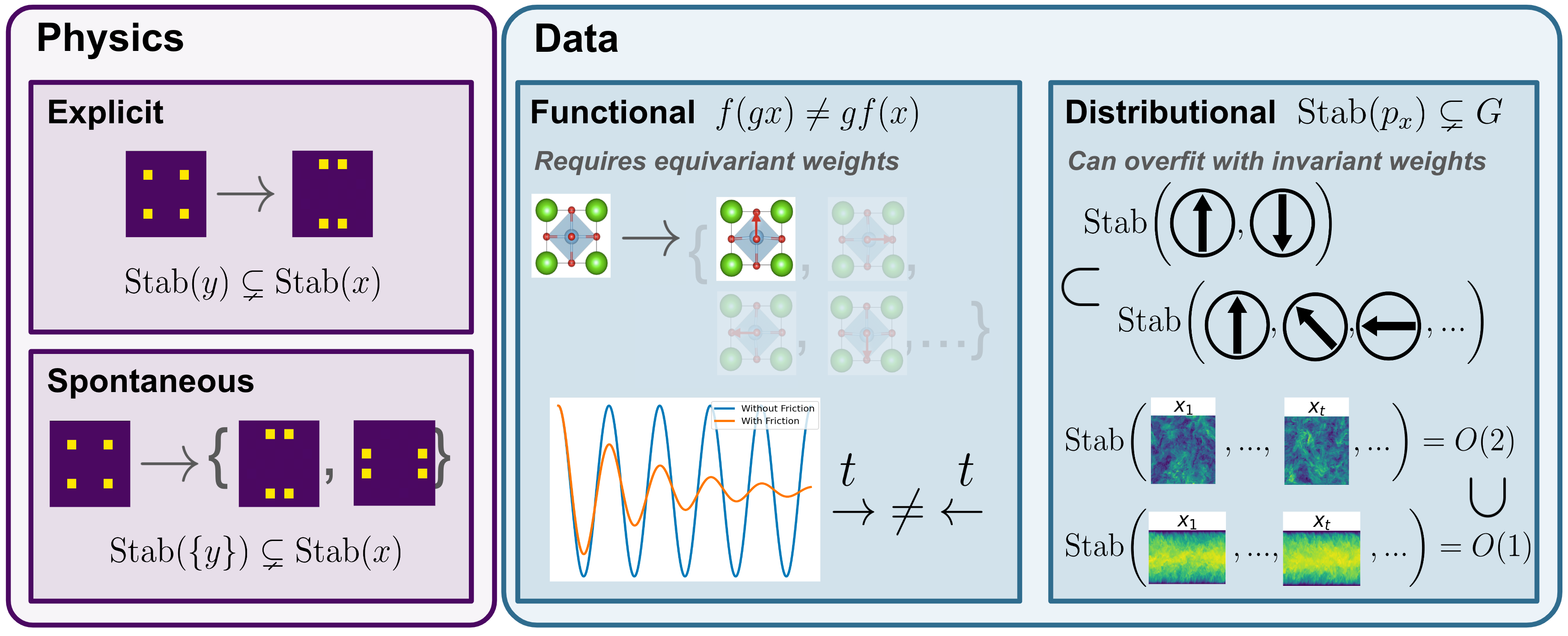}
	\caption{Symmetry breaking can be either explicit, arising when the governing equations are asymmetric, or spontaneous, arising when a symmetric system evolves to a lower symmetry without external influence. It can be also classified as either distributional or functional based on how it manifests in data.}

\label{fig:symmetry_breaking_new}
\end{figure*}

Our method focuses on discovering symmetry-breaking factors from data. The relaxed weights are "analagous" to order parameters used in Landau theory to describe phase transitions \cite{landau1936orderparam}.
Symmetry breaking may arise from: (1) Explicit symmetry breaking: the governing equations of a physical system become variant under the broken symmetry. This could arise from an external force applied to the system or certain boundary conditions. For instance, symmetry-breaking factors, such as gravity, temperature gradient and closed boundaries, break the Euclidean symmetry of a uniform layer of fluid. (2) Spontaneous symmetry breaking: a symmetric system naturally evolves into degenerate lower symmetry states that are energetically favorable without any external symmetry-breaking influence \citep{Castellani2021symmbreak}. For example, in the transition of water to ice, the molecules shift from a random, symmetric arrangement in the liquid state to a well-ordered, less symmetric lattice in the solid state. 


These forms of symmetry breaking manifest themselves in the data, leading us to categorize potential instances of symmetry breaking
as depicted in Figure \ref{fig:symmetry_breaking_new}. 
Assume a function $f: X \mapsto Y$, and $p_{X}:X\mapsto\mathbb{R}$ and $p_{Y}:Y\mapsto\mathbb{R}$ are the probability distributions of the input and output. The expected group $G$ acts on $X$ and $Y$ by $\rho_{X}$ and $\rho_{Y}$. 

(1) Single sample symmetry breaking: 
This could occur when the desired output of the neural network has lower symmetry than the input (e.g. transforming a square to a rectangle).\footnote{When we refer informally to the symmetry of a sample $x$ or a set $S$ we are referring to the stabilizers $\text{Stab}(x) = \{g \in G| g \cdot x = x \}$ and $\text{Stab}(S) =  \{ g \in G | gs \in S, \; \forall s \in S\}$.} We aim to identify the symmetry operations that differ between the input and output or the ``missing information'' needed to make the input and output symmetrically compatible. (See \cite{smidt2021finding} for more discussion.). Formally, $\exists \; x \in X, \; \text{Stab}(f(x)) \subsetneq \text{Stab}(x)$. Note that this is a special case of (3) below.

(2) Distributional symmetry breaking:  In many ideal physical contexts, the entire sample space exhibits symmetry, but a given dataset may or may not. For example, 
the entire sample space of isotropic flow will be closed under rotations because of the axial symmetry of the flow. Nonetheless, certain boundary conditions and external forces may break this rotational symmetry in a given data distribution $p_X$ to varying degrees. Formally, this is stated as $\exists x \in X, \; p_{X}(x) \neq p_{X}(\rho_{X}(g)x)$.\footnote{Defining $\text{Stab}(p_X)$ as $\{g \in G: gp_X = p_X\}$, distributional symmetry breaking can also be stated as $\text{Stab}(p_X) \subsetneq G$.} Distributional symmetry breaking can be detected by an equivariant model with invariant weights. See Appendix \ref{app:fwdpass} for more details and examples. 

(3) Functional symmetry breaking: the mapping between inputs and outputs is not fully equivariant. For instance, a model with time-reversal equivariance, designed to forecast future states based on historical observations, might have trouble accurately predicting dynamics without time-reversal symmetry, such as processes with increasing entropy \cite{jucha2014time}. Formally, $\exists x \in X \;\text{and}\; \exists g \in G, \; f( \rho_{X}(g)(x)) \neq \rho_{Y}(g) f(x)$. Functional symmetry breaking requires equivariant weights to describe. 

\vspace{-5pt}
\subsection{Regular Group Convolution}
Group convolution \citep{Cohen2016Group} achieves equivariance by sharing weights across all group elements. CNNs, for example, achieve shift invariance by implementing group convolution over the group $G = \mathbb{Z}^2$ of discrete translations.

\paragraph{Lifting Convolution}
The first layer of a $G$-convolutional network typically lifts the input to a function on $G$. It maps the input function $f_0$ on $\mathbb{Z}^3$ to functions on a group $G=(\mathbb{Z}^3,+) \rtimes H$, where $H$ is a group acting on $\mathbb{Z}^3$ that we desire equivariance with respect to, such as $C_4$. 
The lifting convolution is the same as conventional convolution with a stack of filter banks transformed by the subgroup $H$,
\begin{equation}
\begin{aligned}
(f_0 \star \psi) (\mathbf{x}, h) 
 = \sum_{\mathbf{y} \in \mathbb{Z}^3}  f_0(\mathbf{y})\psi(h^{-1}(\mathbf{y} -\mathbf{x}))
\end{aligned}
\vspace{-10pt}
\end{equation}
\vspace{-5pt}
\paragraph{Group Convolution}
After the lifting convolution layer, both the filter and input are now functions on $G=(\mathbb{Z}^3,+) \rtimes H$.  
A $G$-equivariant group convolution then takes as input a $c_{\mathrm{in}}$-dimensional feature map $f_1 \colon G \to \mathbb{R}^{c_{\mathrm{in}}}$ and convolves it with kernel $\Psi \colon G \to \mathbb{R}^{c_\mathrm{out} \times c_\mathrm{in}}$ over a group $G$,
\begin{equation}
\begin{aligned}
(f_1 \star \Psi) (\mathbf{x}, h)
     = \sum_{\mathbf{y} \in \mathbb{Z}^3} \sum_{h' \in H}  f_1(\mathbf{y}, h') \Psi(h^{-1}(\mathbf{y-x}),h^{-1}h')\label{gconv}
\end{aligned}
\end{equation}
The last layer usually averages over the $h$-axis and outputs a function on $\mathbb{Z}^3$. 
\vspace{-5pt}
\subsection{Relaxed Group Convolution}
\citet{wang2022approximately} defines relaxed lifting convolution and relaxed group convolution as below. 

\paragraph{Relaxed Lifting Convolution}
Unlike a strictly equivariant layer, the relaxed lifting convolution employs additional trainable weights $w_l(h)$ and may use several filters $\{\psi_l\}_{l=1}^L$ instead of one shared filter.   
\begin{equation}
\begin{aligned}
(f_0 \tilde{\star} \psi) (\mathbf{x}, h) 
 = \sum_{\mathbf{y} \in \mathbb{Z}^3}  f(\mathbf{y})\sum_{l=1}^L w_l(h) \psi_l(h^{-1}(\mathbf{y-x}))
\end{aligned}
\end{equation}

\vspace{-5pt}
\paragraph{Relaxed Group Convolution} The group convolution layer is relaxed in the same way. 
\begin{equation}
\begin{aligned}
&(f_1 \tilde{\star} \Psi) (\mathbf{x}, h) = \\
&\sum_{\mathbf{y} \in \mathbb{Z}^3} \sum_{h' \in H} f_1(\mathbf{y}, h') \sum_{l=1}^L w_l(h) \Psi_l(h^{-1}(\mathbf{y-x}), h^{-1}h')\label{rgconv}
\end{aligned}
\end{equation}
Since the weights $\{w_l(h)\}$ can vary across group element $h$, this can break the strict weight-sharing scheme\footnote{It is not necessary for the weights $w_l(h)$ to be the same for all $l$ or for all layers for equivariance to hold.}. The relaxed weights $\{w_l(h)\}$ in both the relaxed lifting and relaxed group convolution transform as the regular representation of $H$ and are the equivariant weights defined in Section \ref{def:equivariance}. The regular representation contains all irreducible representations (irreps) of $H$ as subrepresentations, and thus the relaxed weights can express symmetry breaking corresponding to any irrep. We will show that the weights $\{w_l(h)\}$ with equal initialization learn to be unequal to break the strict equivariance and adapt to the symmetries of the data when data exhibits broken symmetry.

\section{Theoretical Analysis}
\subsection{Learning symmetry breaking with loss gradients}
In the relaxed group convolution, the initial relaxed (equivariant) weights $\{w_l(h)\}$ in each layer are set to be the same for all $h$, ensuring that the model exhibits equivariance prior to being trained. In this section, we prove that these relaxed weights only deviate from being equal when the symmetries of the input and the output are lower than that of the model.
 
\begin{proposition}\label{main_prop}
Consider a relaxed group convolutional neural network $f$ where the relaxed weights in each layer are initialized to be identical across group elements to maintain $G$-equivariance. If $f$ is trained to map an input $x$ to output $y$, the relaxed weights will change during training such that $f$ is equivariant to $\text{Stab}(x) \cap \text{Stab}(y)$, which is the intersection of the stabilizers of the input and the output.   
\end{proposition}
\vspace{-5pt}
In particular, if $\text{Stab}(x) \cap \text{Stab}(y) < G$, then $w_l(h)$ will not be constant in $h$ due to the gradients of the loss. The proof can be found in Appendix \ref{app:theory}. This useful property allows us to discover symmetry or identify symmetry-breaking factors in data because the relaxed weights tell us whether a transformation $h$ stabilizes the input or output. We provide a clear illustration of this through a simple 2D example. 

\begin{figure}[htb!]
	\centering
	\includegraphics[width=0.5\textwidth]{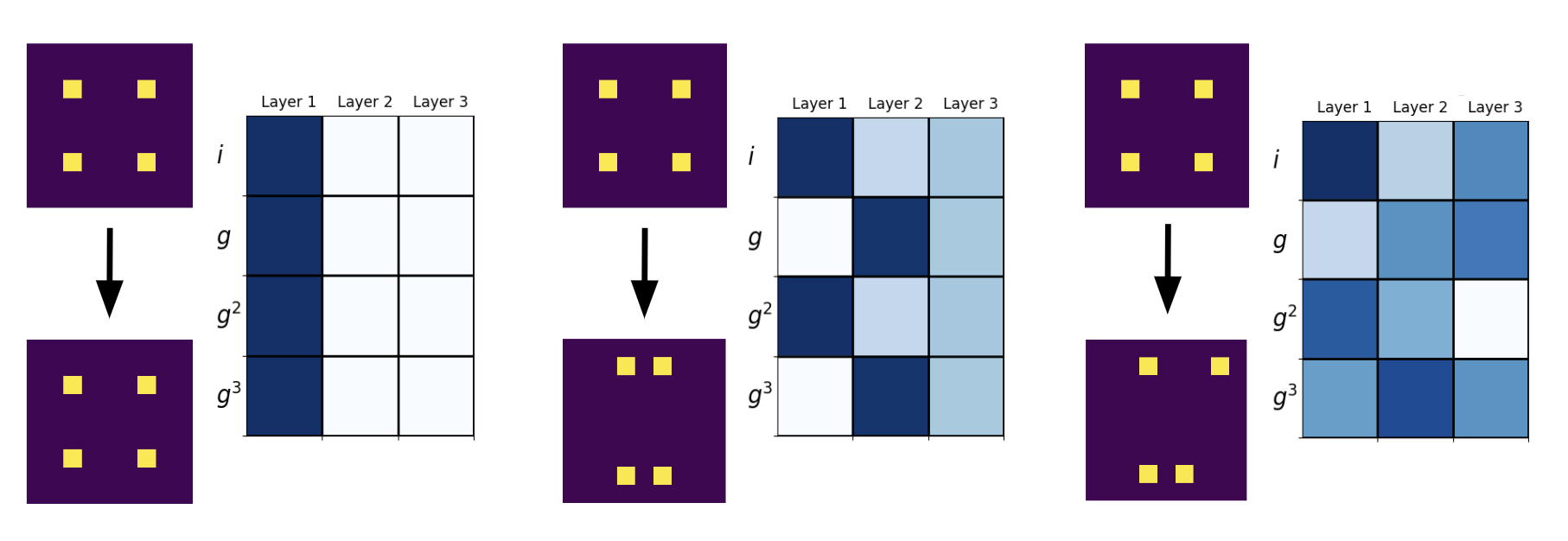}
	\caption{Visualization of tasks and corresponding relaxed weights after training. A 3-layer $C_4$-relaxed group convolution network with $L=1$ is trained to perform the following three tasks: 1) map a square to a square; 2) deform a square into a rectangle; 3) map a square to a non-symmetric object.}
	\label{fig:2d_example}
\end{figure}

As shown in Figure \ref{fig:2d_example}, in the first task where the output is a square with $C_4$ symmetry\footnote{We ignore reflections in this example.}
, relaxed weights across all layers remain equal throughout training. For the second task, where the output is a rectangle exhibiting $C_2$ symmetry, the relaxed weights learn to be different. However, the weights corresponding to the group elements $i$ and $g^2$ are the same, as are the weights for $g$ and $g^3$. This reflects the symmetry of the rectangle as the model becomes equivariant to $C_2$ (180 degree rotations). In the final task, where the output lacks any symmetries, the relaxed weights diverge entirely for the four group elements, thereby fully breaking the model's equivariance.

\vspace{-5pt}
\subsection{Convergence of relaxed weights using mini-batch gradient descent}
Proposition \ref{main_prop} also indicates that relaxed group convolution can discover symmetries that lie in the data distribution using full batch gradient descent, as the entire dataset can be viewed as a single sample in this case. However, when dealing with high-dimensional data, it often becomes necessary to use mini-batch gradient descent, where mini-batches may not represent the symmetries of the entire dataset. Thus, we have the following proposition. 

\begin{proposition}\label{minibatch}
Denote $\mu$ the step size of the gradient 
descent, $M$ the number of mini-batches in each epoch, and $t$ the number of epochs respectively. Assume that the loss 
function $L$ is convex and has $\delta$-Lipschitz continuous gradients. Consider $\bm{w}^*$ the optimal relaxed weights that can be reached by full batch gradient descent, with $K$ representing the gradient noise variance at $\bm{w}^*$. Then the starting relaxed weights $\bm{w}^t_0$ at the $t$-th epoch satisfy
\begin{align*}
    E[\|\bm{w}^t_0 - \bm{w}^*\|] \leq & \left(1 - \frac{\mu^2\delta^2 M^2}{2\mu^2\delta^2 M^2-1}\right)^tE[\|\bm{w}^0_0 - \bm{w}^*\|] \\
    & + \frac{\mu^4\delta^2 M^4 K}{1-2\mu^2\delta^2 M^2}
\end{align*}
\end{proposition}
\vspace{-5pt}
The convergence rate $1 - \frac{\mu^2\delta^2 M^2}{2\mu^2\delta^2 M^2-1}$ is bounded by $\frac{1}{2}$ and it will descrease as the batch size increases (i.e. $M$ becomes smaller). Given a sufficient number of epochs (i.e. when $t$ is large enough), the relaxed weights learned by mini-batch gradient descent will converge to the optimal relaxed weights learned by the full batch gradient descent. Therefore, training relaxed group convolution with mini-batches still allows for the post-training relaxed weights to uncover the symmetries in the dataset.

\vspace{-5pt}
\subsection{Decomposition of relaxed weights into irreps}
By projecting the relaxed weights onto the irreps of the group, similar to calculating its Fourier components, one can identify the type of geometric symmetry breaking and which symmetries are preserved or broken. This is particularly useful for larger groups.

Consider the relaxed weights from a neural network layer illustrated in Figure \ref{fig:2d_example}, projected onto $C_4$'s four one-dimensional irreps. For the first task, only the Fourier component for the trivial representation is non-zero. In the second task, Fourier components for both the trivial and the sign representation are non-zero, suggesting that either 90 or 180-degree rotational symmetries are broken. Since the remaining irreps are zero, we can conclude the output is still invariant under 180-degree rotations. In the third task, all Fourier components are non-zero. Thus, due to the property that the relaxed group convolution reliably preserves the highest level of equivariance that is consistent with data, it can discover symmetry or symmetry breaking in data. 

\section{Related Work}
\vspace{-3pt}
\subsection{Approximate Symmetry}
Equivariant deep learning models have excelled in image analysis\citep{chidester2018rotation, weiler2019e2cnn, kondor2018generalization, bao2019equivariant, worrall2017harmonic, finzi2020generalizing, Ghosh19Scale}, and their applications have also expanded to physical systems due to the deep relationship between physical symmetries and the principles of physics \citep{geiger2022e3nn, wang2020incorporating, otto2023unified, Anderson2019Cormorant, satorras2021n, liu2022spherical, zitnick2022spherical, schutt2017schnet, passaro2023reducing}. 
However, real-world data rarely conforms to strict mathematical symmetries, due to noise, missing values, or symmetry-breaking features in the underlying physical system. Thus, recent works have tried to relax the strict symmetry constraints imposed on the equivariant networks. 
\citet{Elsayed2020Revisiting} first showed that relaxing strict spatial weight sharing in 2D convolution can improve image classification accuracy.
\citet{wang2022approximately} generalized this idea to arbitrary groups and proposed relaxed group convolution, which is biased towards preserving symmetry but is not strictly constrained to do so. 
\citet{petrache2023approximation} further provides a theoretical study of how the data equivariance error and the model equivariance error affect the models' generalization abilities. 
In this paper, we further apply relaxed group convolution to various physical systems cases and reveal its ability in symmetry discovery. Alternatively, \citet{finzi2021residual} proposed a mechanism that sums equivariant and non-equivariant layers but it cannot handle high-dimensional data due to the number of weights in the fully-connected layers.
\citet{van2023learning} proposed a method to empirically learn the amount of layer-wise equivariance through approximate Bayesian model selection but it does not have theoretical guarantees of learning the correct amount of symmetry from the data.
\citet{mcneela2023almost} proposed the Lie
algebra convolution that relaxes the strict equivariance constraints in Lie group convolutions but it cannot be used to find the symmetry-breaking factors. 
\vspace{-6pt}
\subsection{Symmetry Discovery} \label{related_work_baselines}
Our method can also perform symmetry discovery as we can uncover hidden symmetries in data by creating relaxed group convolution layers of the largest possible group.  This differs from most existing methods for finding symmetries, which often require intricate architecture and careful tuning. For example, \citet{desai2022symmetry, yang2023generative, yang2023latent} designed GAN-based architectures where the generator is trained to generate group transformations that stabilize the data distribution. \citet{zhou2021metalearning} factorized the weight matrix into a symmetry matrix, which learns the weight-sharing pattern from the data, and a separate vector for filter parameters. These two parts are trained separately with MAML, which is an optimization-based meta-learning method \cite{finn2017model}. Furthermore, \citet{dehmamy2021automatic} proposed L-conv, a novel architecture that can learn the Lie algebra basis and automatically discover symmetries from data. Unlike these complex models and training approaches, our method simplifies the process. By utilizing the equivariance property of gradients, the relaxed weights in a single relaxed group convolution layer can discover symmetry breaking with minimal training, either on a single symmetric
sample for one iteration or on a symmetric dataset for several epochs.



\section{Experiments}
\subsection{Discovering Symmetry Breaking Factors in Phase Transitions of Crystal Structures}\label{exp:crystal}
\paragraph{Phase Transition}
Modeling a phase transition from a high symmetry (like octahedral) to a lower symmetry is a common topic of interest in materials science \cite{onuki2002phase}. This change can be understood as a transformation in the arrangement or orientation of atoms within a crystal lattice. For instance, perovskite crystal structures consist of connected octahedral motifs that under certain conditions distort and give rise to technologically relevant material properties. 

\begin{figure}[htb!]
\centering
\includegraphics[width=0.5\textwidth]{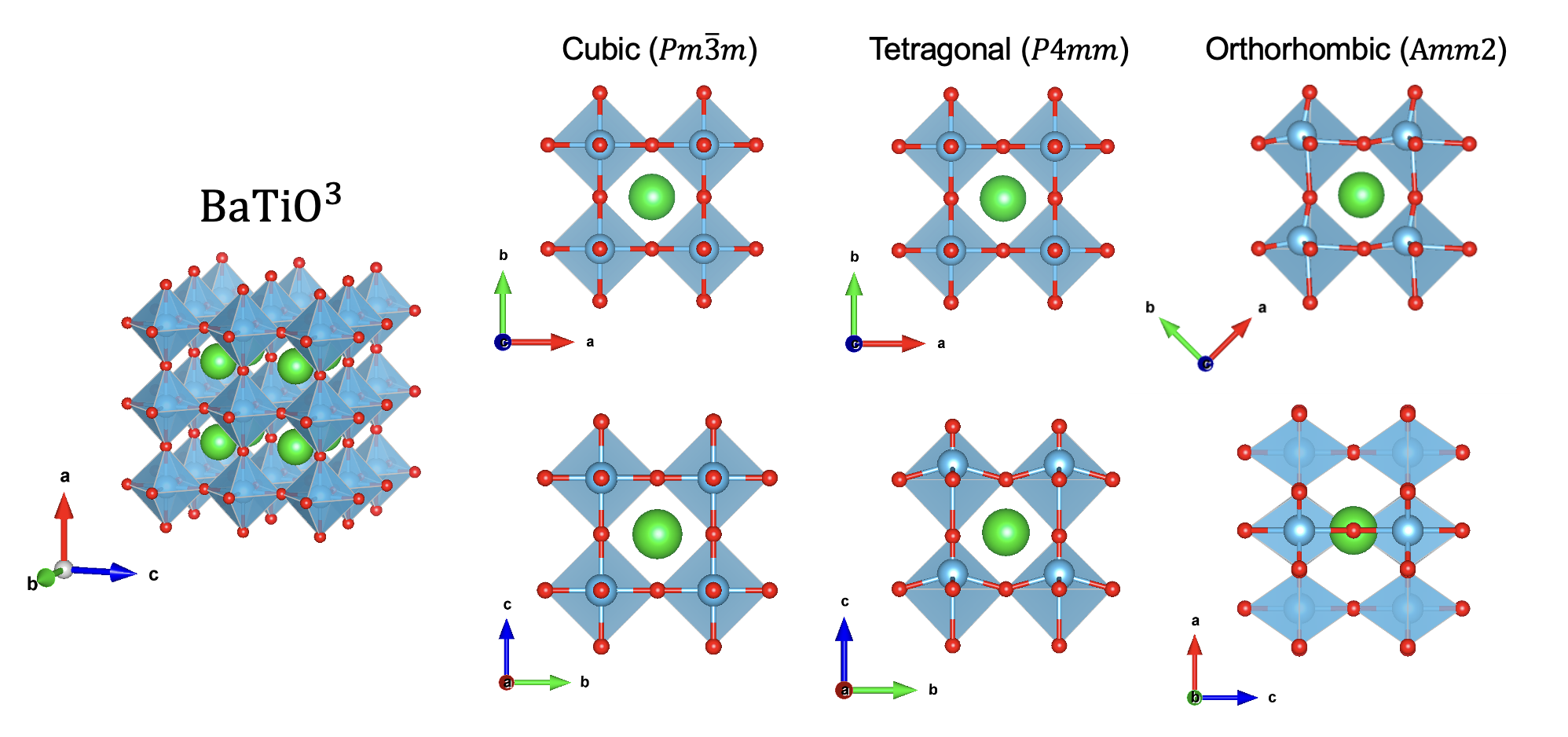}
\vspace{-5pt}
\caption{Visualization of $BaTiO_3$: As temperature decreases, it undergoes a series of symmetry-breaking phase transitions, transitioning from a cubic structure to a tetragonal phase, and eventually to an orthorhombic form.}
\label{fig:batio3}
\end{figure}

To illustrate, consider 
the perovskite of Barium Titanate ($BaTiO_3$), as shown in Figure \ref{fig:batio3}. At high temperatures,  $BaTiO_3$ has a cubic perovskite structure with the $Ti$ ion at the center of the octahedron. As one progressively cools $BaTiO_3$, it undergoes a series of symmetry-breaking phase transitions. Initially, around 120°C, there is a shift from the cubic phase to a tetragonal phase, where the Ti ion is displaced from its central position. The space group of the tetragonal system in our example is $P4mm$ and the point group is $C_{4v}$.
This is followed by a transition to an orthorhombic system at about -90°C, the space group and the point group of which are $Amm2$ and $C_{2v}$ respectively. 
\vspace{-5pt}
\paragraph{Model Design}
We build relaxed octahedral group convolution layers to model the phase transition of crystal structures. The octahedral group, denoted $O_h$, contains 48 elements. It is a finite subgroup of $O(3)$ that describes all the symmetries of a regular octahedron or a cube. The octahedral group is compatible with the cubic lattice and the highest symmetry group of crystals \citep{nabarro1947dislocations, van19922p, woodward1997octahedral}. 

\begin{figure}[htb!]
	\centering
	\includegraphics[width=0.48\textwidth]{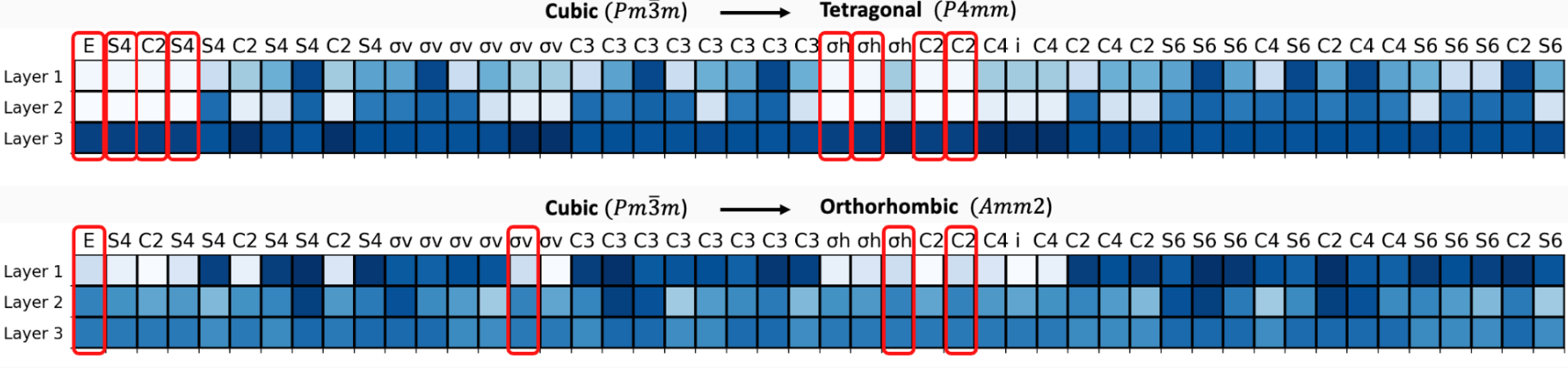}
	\caption{Visualization of the relaxed weights of two 3-layer relaxed octahedral group convolution networks trained to map from the cubic system to the tetragonal system and the orthorhombic system. The labels on top of columns represent the classes of the octahedral group. It is easier to annotate each set of relaxed weights with the class names rather than the actual transformations. The highlighted relaxed weights correspond to the preserved symmetry operations in these two systems that forms $C_{4v}$ and $C_{2v}$ group respectively.}
	\label{fig:batio3_all_weights}
\end{figure}

We train a 3-layer relaxed $O_h$ group convolutional network with a single filter basis to map from the cubic phase to a tetragonal phase, and from the cubic phase to an orthorhombic system. Since both the input and the network have $O_h$ symmetry and the output is lower than $O_h$, the post-training relaxed weights reveal the symmetry-breaking factors in these two phase transitions. 
\vspace{-5pt}
\paragraph{Dataset}
We use the Cartesian coordinates of $BaTiO_3$ in cubic, tetragonal, and orthorhombic phases from the Material Project \cite{jain2013commentary}. We use 3D tensors to represent these systems where the voxels corresponding to atoms are non-zero. 
We only focus on the point group symmetry and do not consider translation symmetry breaking in this experiment. Instead, we will explore translation symmetry in the channel flow experiment. 
\vspace{-5pt}
\paragraph{Experimental Results}
Figure \ref{fig:batio3_all_weights} visualizes the relaxed weights from the two 3-layer models trained to predict the tetragonal and orthorhombic structures from a cubic system. The highlighted relaxed weights are the same as those of the identity element, corresponding to preserved symmetry operations. Since both the input and the model have $O_h$ symmetry, based on Theorem \ref{main_prop}, the highlighted elements stabilize the output of the networks.  Those two sets of preserved symmetry operations found by the networks form the $C_{4v}$ and $C_{2v}$ groups respectively, which align with the point group of the tetragonal system ($P4mm$) and the point group of the orthorhombic system ($Amm2$) \cite{bradley2010mathematical}. These results demonstrate the capability of relaxed group convolution to automatically adapt to the symmetry of the output. 

We can further investigate the symmetry-breaking factors by decomposing the relaxed weights that transform under regular representation into the 10 irreducible representations (irreps) of $O_h$. The preserved symmetries after phase transitions are the intersection of the stabilizers of all non-zero irreps. More details can be found in Appendix \ref{app:irrep_analysis}. As the asymmetry of mapping in this case is due to spontaneous symmetry breaking, to recover all possible configurations, one would need to take the orbit of degenerate solutions.



\vspace{-3pt}
\subsection{Discovering Isotropy Breaking in Turbulence}\label{exp:isotropy}
\paragraph{Isotropy Breaking} Kolmogorov's Hypothesis \cite{zakharov2012kolmogorov, pope2001turbulent} states that, at sufficiently high Reynolds numbers, the small-scale turbulent motions are statistically isotropic and independent of the large-scale structure. This implies that larger eddies lack rotational symmetry due to factors such as boundary conditions and external forces. As they continue to break into smaller eddies and these eddies become sufficiently small, they begin to display rotational symmetries in their distribution. The hypothesis raises the question of determining the specific scale at which fluid symmetries are restored. 

\begin{figure*}[htb!]
  \begin{minipage}[b]{0.43\textwidth}
   \includegraphics[width=\textwidth]{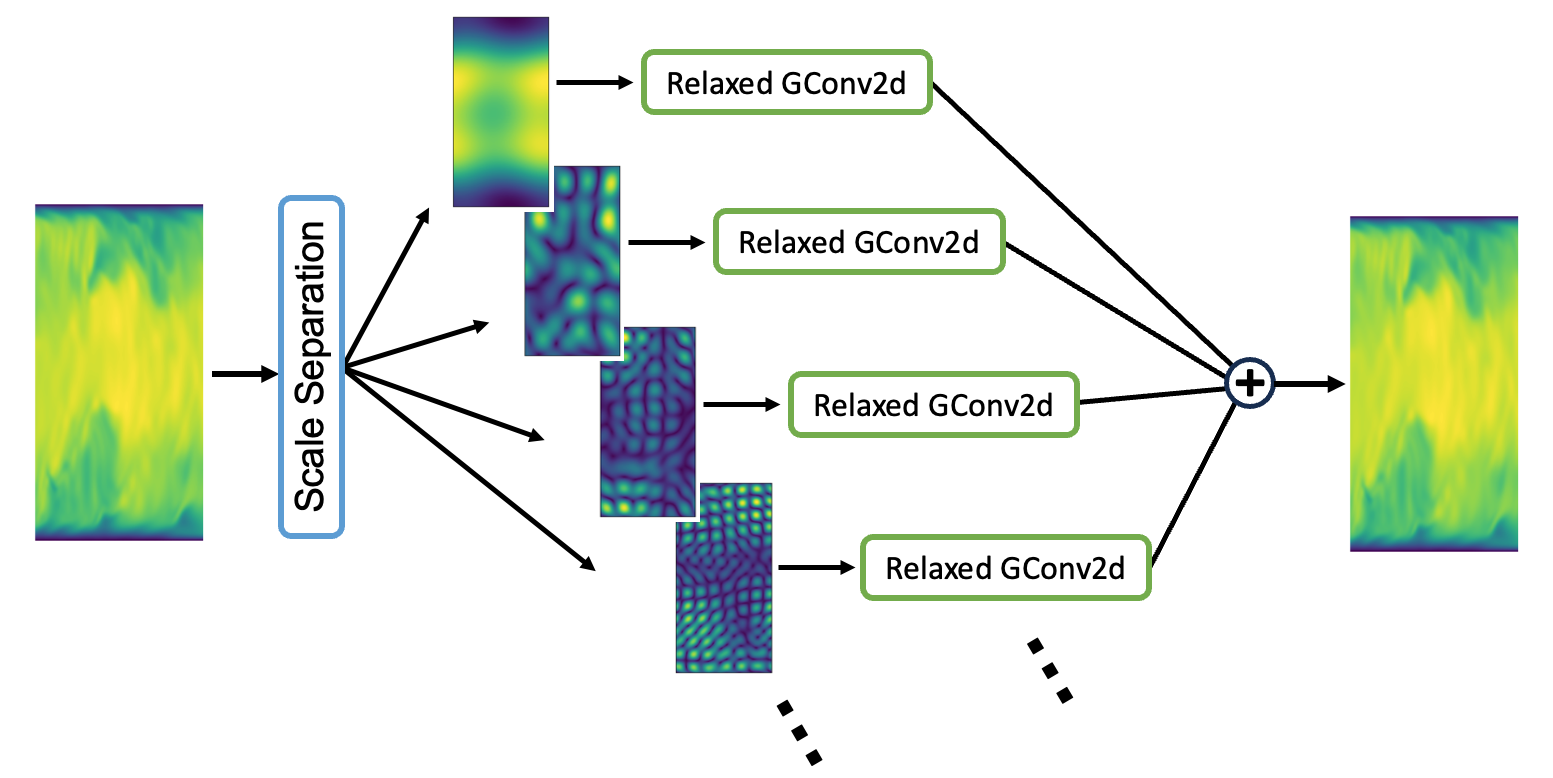}
  \end{minipage} 
   \begin{minipage}[b]{0.25\textwidth}
   \includegraphics[width=\textwidth]{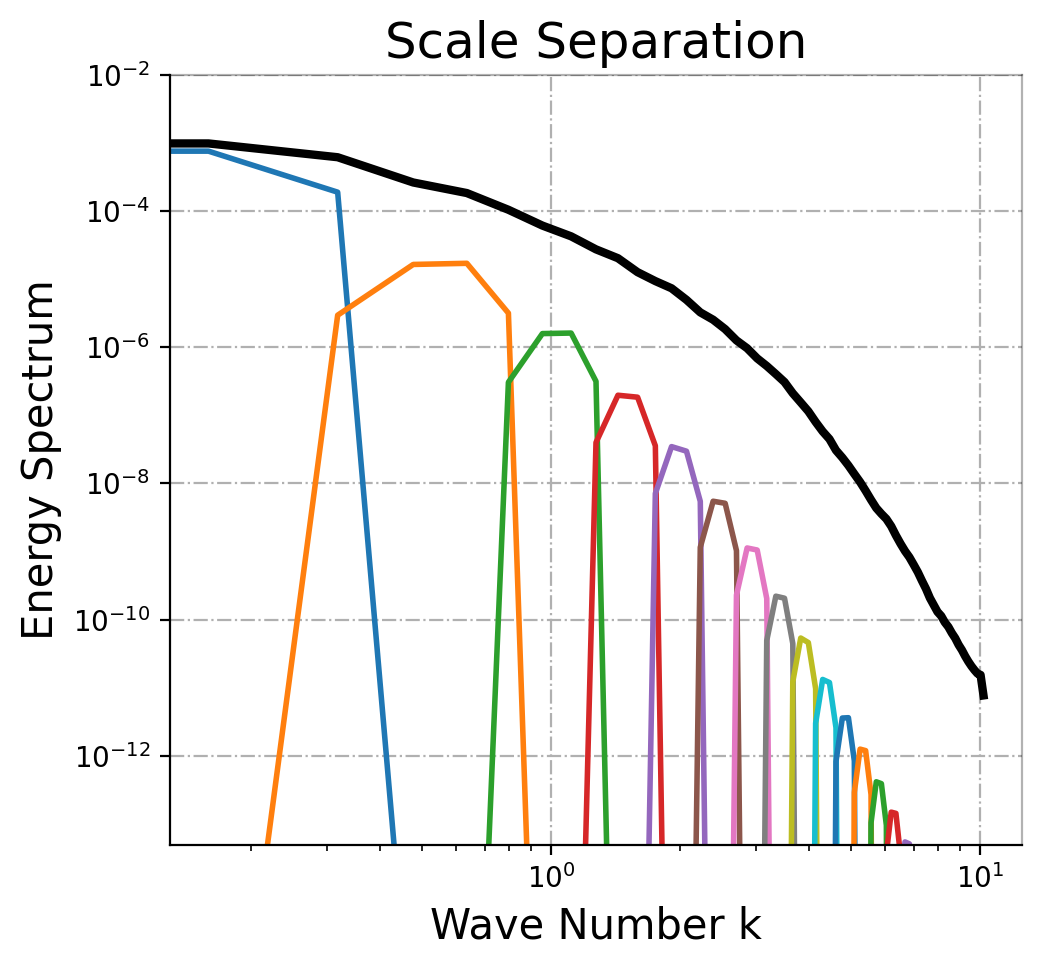}
  \end{minipage} 
  \begin{minipage}[b]{0.3\textwidth}
     \includegraphics[width=\textwidth]{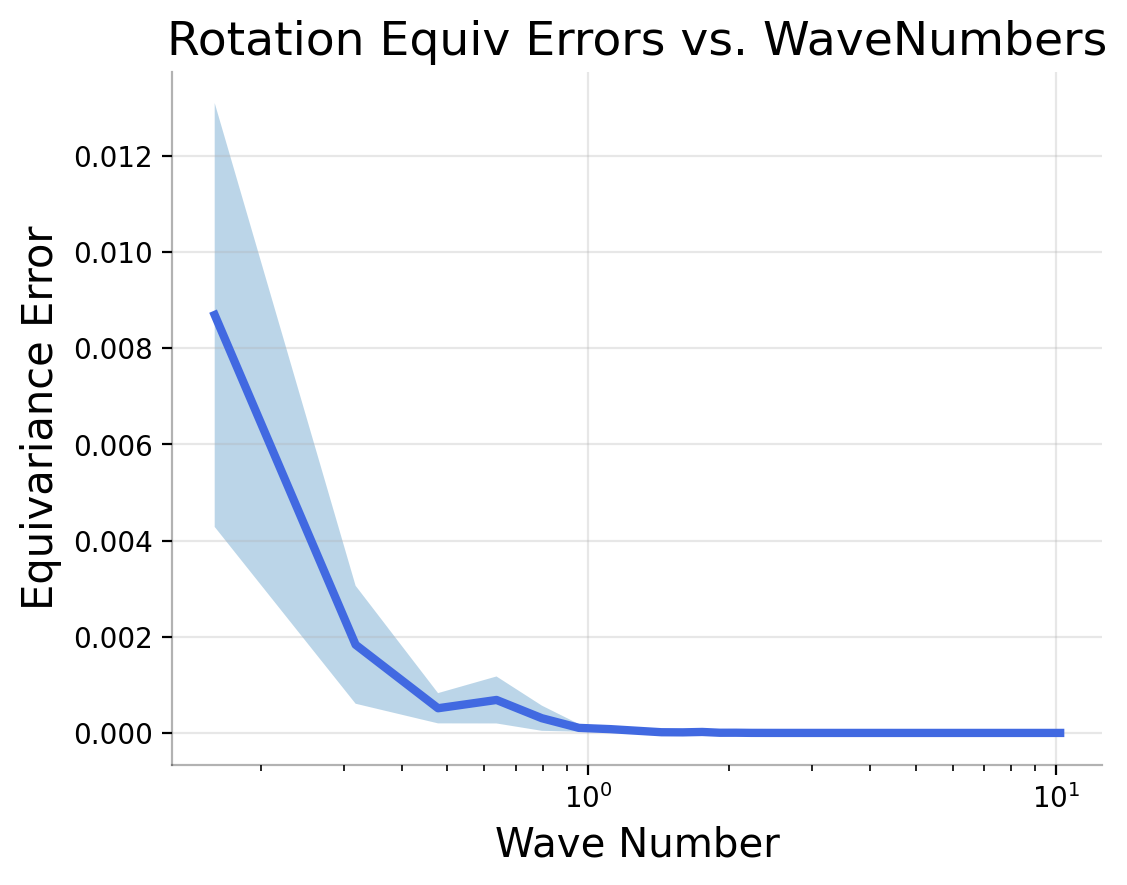}
  \end{minipage} 
\caption{Left: Model for detecting rotational symmetry. This model breaks down the input velocity field into multiple scales using Fourier frequency cutoffs. Each scale is then processed through a distinct relaxed group convolution layer and the sum of the outputs from these layers is trained to reconstruct the input. Middle: Visualization of scale separation in energy spectrum. The black line represents the energy spectrum of the original velocity fields, while the other colored lines correspond to different scales. Right: Equivariance errors learned by the model. As the wave number gets higher (indicating smaller eddies), the equivariance error tends to decrease towards zero, indicating the isotropy crossover point.}
\label{fig:isotropy}
\end{figure*}
\vspace{-5pt}
\paragraph{Model Design} To identify the symmetries in eddies across various frequencies using relaxed group convolution, it is essential to separate the input velocity fields into different scales. This is achieved by applying different bandpass filters\textemdash applying Fourier transformation, frequency cutoff, and inverse Fourier transformation to each frequency range, as illustrated in the left subfigure of Figure \ref{fig:isotropy}. The middle subfigure in Figure \ref{fig:isotropy} visualizes the scale separation in the energy spectrum. The black line represents the energy spectrum of the original velocity fields, while the other colored lines correspond to different scales. The turbulence energy spectrum is a quantitative representation of the distribution of kinetic energy across different scales or frequencies of turbulence in a fluid. It characterizes how energy in large eddies transformed into small ones. More details about the energy spectrum are in Appendix \ref{app:energy_spectrum}

After scale separation, each scale is then processed through a distinct relaxed group convolution layer.  Since our focus is on discovering symmetry, we aim to make our models and learning tasks as simple as possible for ease of use. The sum of the outputs from these layers is trained to reconstruct the input. By examining the relaxed weights in the layer corresponding to each scale, the degree of symmetry present at each scale can be determined. We define the equivariance error of a relaxed group convolution layer $f: X \mapsto Y$ as follows,
\begin{equation} \label{eqn:equiv_error}
    \|f\|_{\text{EE}} = \frac{1}{L|G|}\sum_{l=1}^L\sum_{g\in G}\|w_{l}(g) - w_{l}(e)\|
\vspace{-5pt}
\end{equation}
where $G$ is the relevant group ($G$ is $C_8$ in this experiment), $L$ is the number of filter banks used in the layer, and $e$ is the identity element in the group $G$. Since $\{w_{l}(g)\}_{g\in G}$ are initialized as equal, $\|f\|_{\text{EE}}$ is zero before training. The bigger $\|f\|_{\text{EE}}$ is after training, the more symmetry breaking there is in the data. Note that the magnitude of relaxed weights is affected by the magnitude of the input. Therefore, the relaxed weights need to be normalized at each scale prior to computing equivariance errors and making comparisons.
\vspace{-5pt}
\paragraph{Dataset}
We use a 2D direct numerical simulation of a turbulent boundary layer flow/channel flow from \citet{gao2023bayesian}, which includes a 2000-step simulation of the velocity field at a resolution of 128 x 256. In this dataset, the fluid flows from left to right, bounded by closed boundaries at the top and bottom, as shown in Figure \ref{fig:isotropy}. 
Note that while individual velocity frames do not exhibit symmetry, the entire data distribution may have symmetry. This differs from the earlier example with crystals. In this experiment, our objective is to uncover the rotational invariance of the distribution of the fluid velocities or, more specifically, to detect the statistical isotropy inherent in turbulence. 
\vspace{-5pt}
\paragraph{Experimental Results}
The model is trained to reconstruct each velocity field of the turbulent flow simulation using mini-batch gradient descent. We employ an 80\%-20\% split for training and validation and use early stopping based on the validation loss. 
After training, the equivariance error for each relaxed group convolution layer is calculated based on Eqn.\eqref{eqn:equiv_error} to obtain the degree of symmetry breaking at different scales. As shown in the right subfigure in Figure \ref{fig:isotropy}, as the wavenumber increases (i.e. eddies become smaller), the equivariance errors decrease and approach zero. This implies that the smaller eddies have perfect rotation symmetry, thus empirically validating the Kolmogorov Hypothesis!

Notably, since we use properties of the gradient to discover symmetry, the choice of hyperparameters and the prediction performance of the models do not influence this symmetry-breaking result.

\subsection{Discovering Homogeneity Breaking in Turbulence}\label{exp:homogeneity}
\begin{figure}[htb!]
  \begin{minipage}[b]{0.33\textwidth}
   \includegraphics[width=\textwidth]{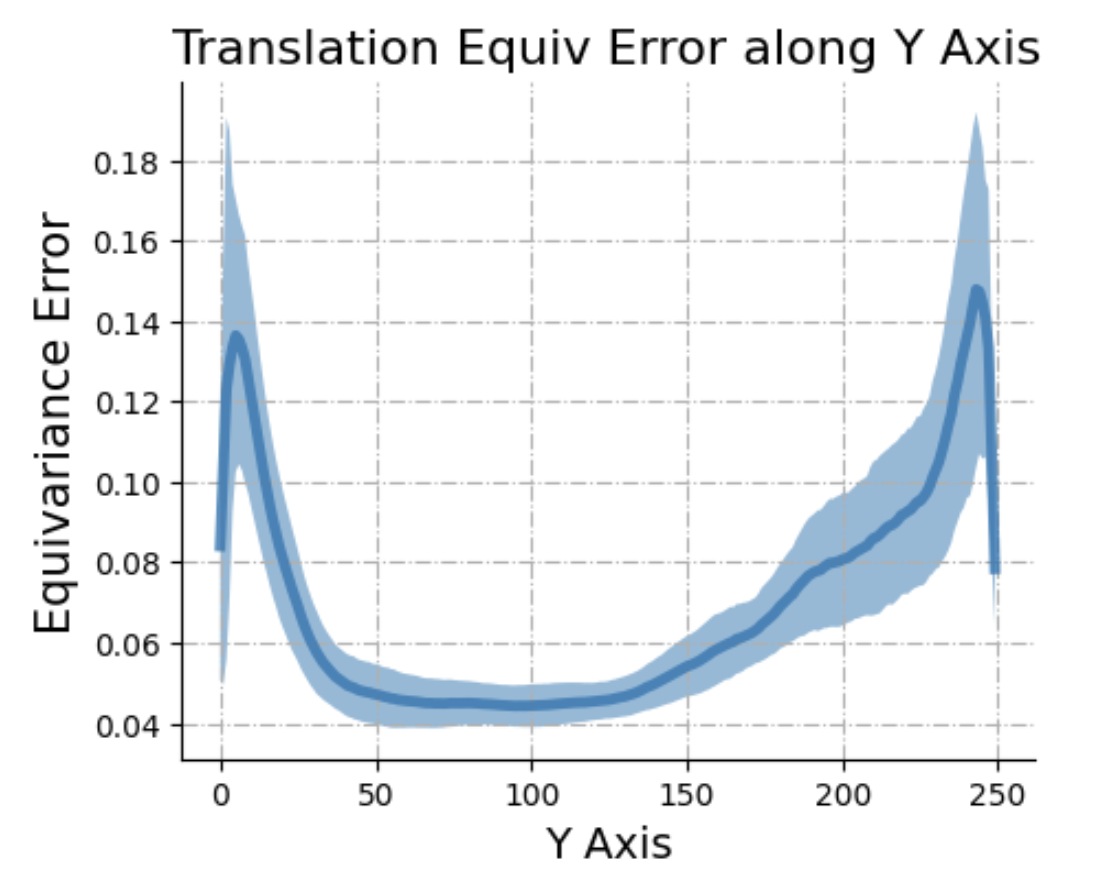}
  \end{minipage} \hspace{5pt}
   \begin{minipage}[b]{0.13\textwidth}
   \includegraphics[width=\textwidth]{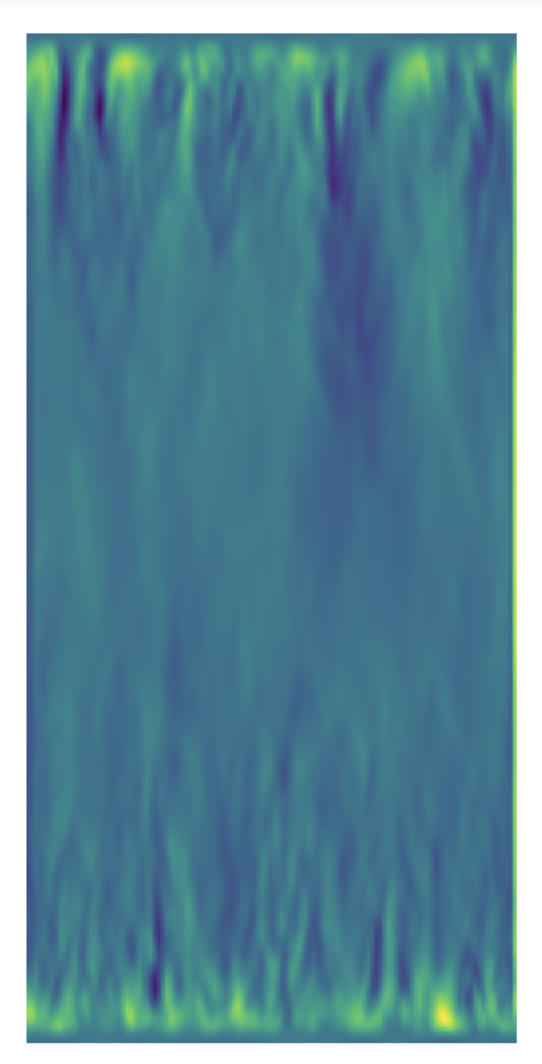}
  \end{minipage}
  \caption{Left: Translation equivariance error along the Y-axis. The level of symmetry breaking increases near the top and bottom boundaries. Right: The visualization of translation relaxed weights. It shows increased variation in relaxed weights as they approach the boundary areas.}
  \label{fig:translation}
\end{figure}
\vspace{-5pt}
\paragraph{Homogeneity Breaking} The turbulence in idealized scenarios is also homogeneous, which means the time-averaged properties of the flow are uniform and independent of position \cite{wang2020incorporating, batchelor1953theory}. This implies that the distribution of homogeneous turbulence exhibits translation symmetry. Nevertheless, factors like noise and boundary conditions may break this translation symmetry to varying degrees. We will show that the positions of translation symmetry breaking can also be uncovered by the relaxed translation group convolution. 
\vspace{-5pt}
\paragraph{Model Design} 
The relaxed translation group convolution layer relaxes the translation symmetry by assigning an additional trainable parameter to each spatial position of the input such that every input position does not necessarily share the same convolution kernel. We use the same channel flow dataset and simply train a 3-layer relaxed translation convolution neural network to map each velocity field to itself. The resolution of the simulation is 256$\times$128. 
\vspace{-5pt}
\paragraph{Experimental Results}
In Figure \ref{fig:translation}, the right subfigure is the heatmap of translation relaxed weights of the first layer of the model. We can see the increased variation in relaxed weights as they approach the top and bottom boundaries. We can calculate the equivariance error along the vertical or Y-axis using Eqn. \eqref{eqn:equiv_error}. We can then find out how much translation symmetry is broken at different heights. The left subfigure in Figure \ref{fig:translation} shows the translation equivariance error along the Y-axis. We can see that the level of symmetry breaking increases near the top and bottom boundaries.

During simulations, perturbation is introduced into the boundaries of simulations to replicate real-world disturbances, which primarily contribute to turbulence.  Thus, the translation symmetry or the homogeneity of the fluid is expected to mostly be broken around the boundary regions. This aligns precisely with our findings from the weights of the relaxed translation group convolution networks. The reason why Figure \ref{fig:translation} is not perfectly symmetric along the y-axis is that the simulation is not long enough to encompass all possible samples. 

Notably, this experiment further demonstrates relaxed group convolution can parameterized to be sensitive not solely to global symmetry breaking but also local symmetry breaking.

\vspace{-3pt}
\subsection{Discovering Time Reversal Symmetry Breaking}
\paragraph{Time Reversal Symmetry Breaking} 
Time reversal symmetry \cite{lamb1998time} holds for physical laws that remain unchanged when the direction of time is reversed. This symmetry is a key aspect in theoretical physics and has profound implications in various fields such as quantum mechanics, statistical mechanics, and thermodynamics. But not all physical processes exhibit time-reversal symmetry, such as processes that involve entropy increase \cite{luke1998time, jucha2014time}. 

\begin{figure}[htb!]
  \begin{minipage}[b]{0.235\textwidth}
   \includegraphics[width=\textwidth]{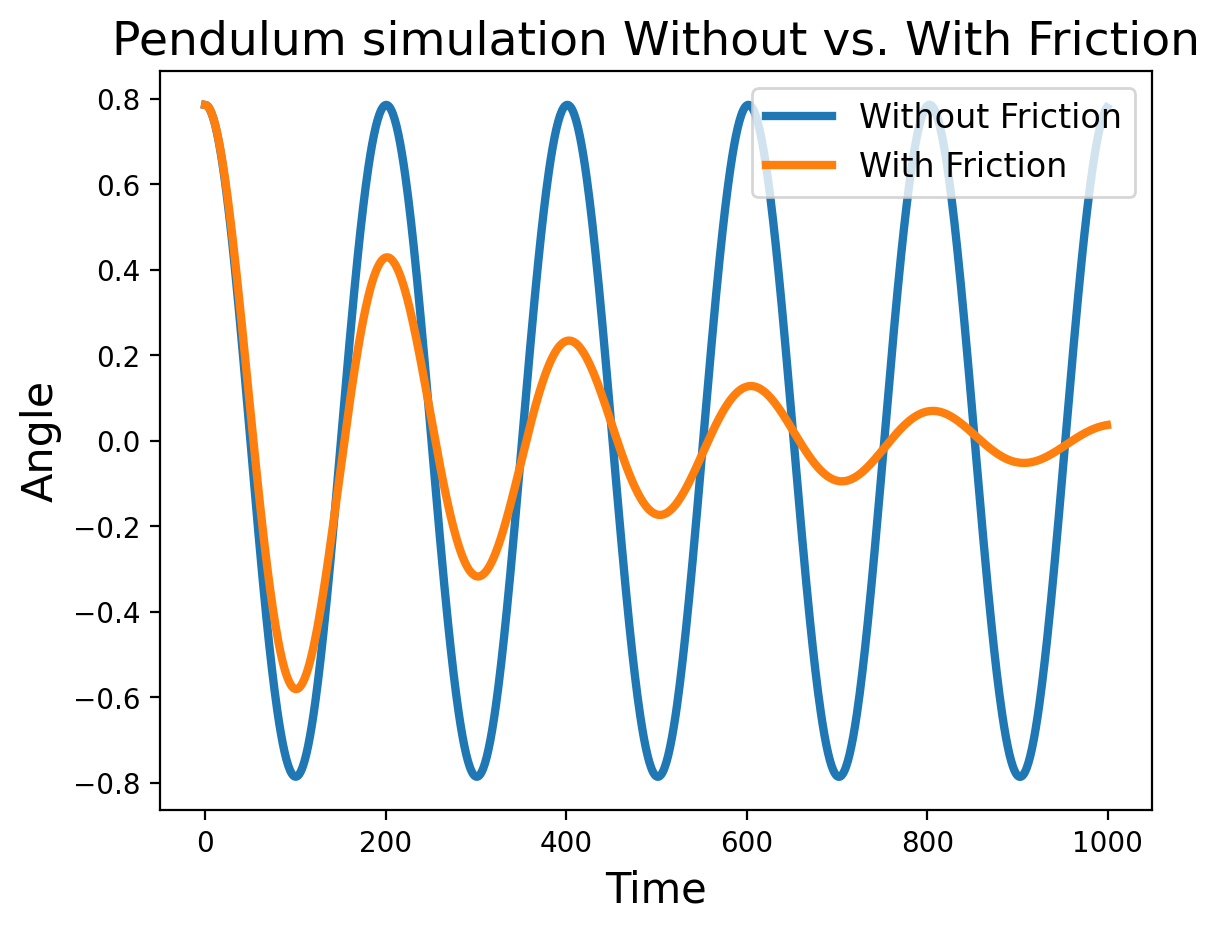}
  \end{minipage}
   \begin{minipage}[b]{0.235\textwidth}
   \includegraphics[width=\textwidth]{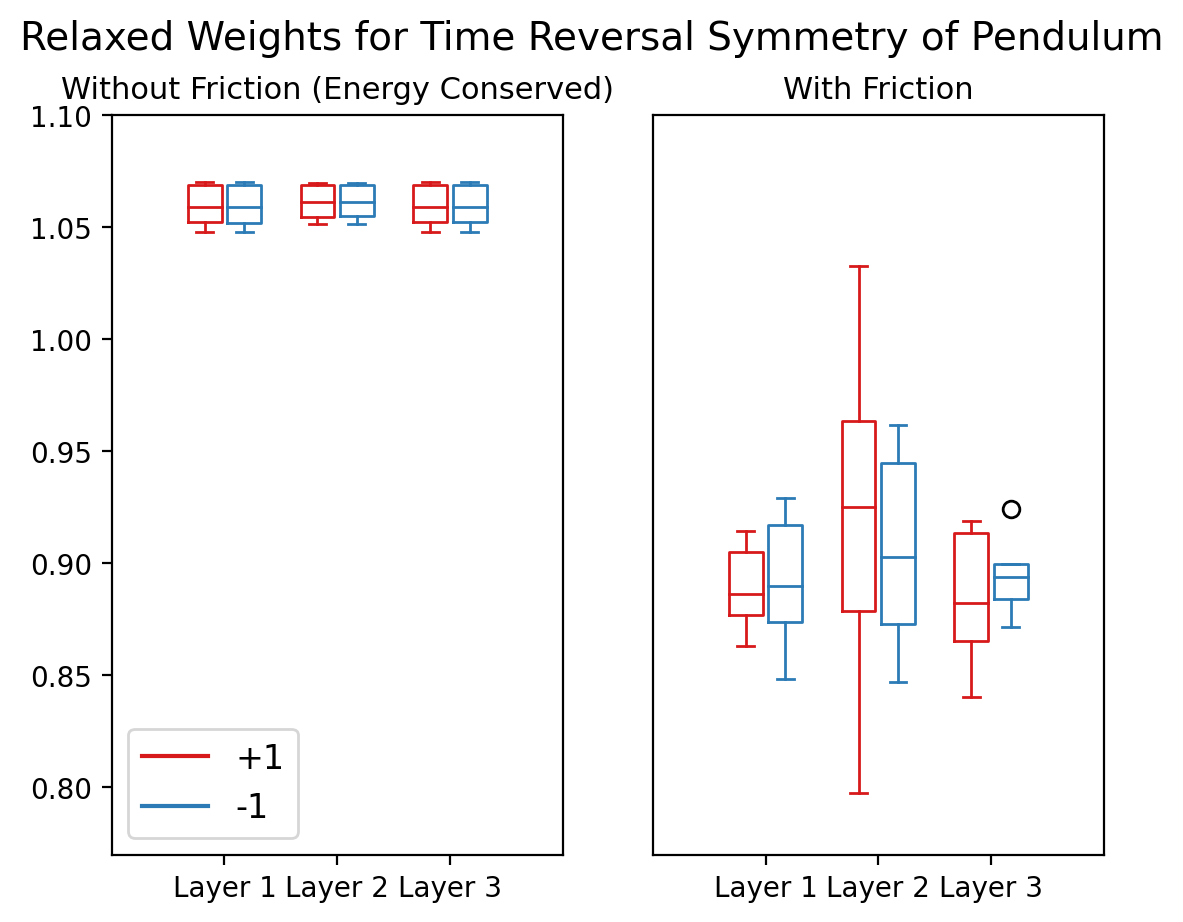}
  \end{minipage}
  \caption{Left: Visualization of the pendulum simulations without and with frictions. Right: the relaxed weights for time reflection from models trained on simulations without and with frictions.}
  \label{fig:time_reversal}
\end{figure}

A straightforward example is a pendulum's motion. Figure \ref{fig:time_reversal} visualizes the pendulum simulations without and with friction. In an ideal, frictionless environment, a pendulum swinging back and forth demonstrates time-reversal symmetry. Its motion looks natural and consistent whether viewed normally or in reverse. However, when friction is introduced, this symmetry breaks. Friction causes the pendulum to lose energy and eventually stop. This process would not look the same if time were reversed. This illustrates how time reversal symmetry can be present in idealized systems but is often broken in real-world scenarios with dissipative forces like friction.
\vspace{-5pt}
\paragraph{Model Design}
Since we want to discover time-reversal symmetry, we have developed a one-dimensional relaxed group convolution model that maintains equivariance with the reflection $\{+1, -1\}$ w.r.t the time dimension. Thus, every layer only contains two scalar relaxed weights if we only use a single filter basis (i.e. $L=1$). If these two weights differ after training, it would suggest a break in the symmetry of time reversibility. To test this, we employed a three-layer network adapted for relaxed time reflection group convolution, aiming to predict the pendulum's angle variations over time. Both the input and the output contain multiple frames. 
\vspace{-5pt}
\paragraph{Dataset}
Relaxed time reflection group convolution networks were trained using simulations of pendulum systems initialized at various initial angles. Each simulation trajectory is sliced and follows the standard preprocessing procedure for time series predictions \cite{benidis2022deep}. 
\vspace{-5pt}
\paragraph{Experimental Results}
We compare the relaxed time reflection weights from models trained on the simulations without and with frictions. As shown in Figure \ref{fig:time_reversal}, we can see the relaxed weights stay the same across all three layers when the model is trained on frictionless simulations. In contrast, these weights vary when the models are trained on simulations with frictions. These boxplots are based on 10 runs with different random seeds. This result demonstrates the capability of relaxed group convolution in identifying time-reversal symmetry.

\subsection{Super-resolution of 3D Turbulence}
We also evaluate the performance of regular, group equivariant, and relaxed group equivariant layers built into this architecture in the tasks of upscaling 3d channel flow and isotropic turbulence. We observe that imposing relaxed equivariance consistently yields better prediction performance. More importantly, Figure \ref{fig:vis_relaxedweights_flow} in the appendix shows that for isotropic flow, the relaxed weights remain nearly constant across group elements, aligning with its rotational symmetries. Channel flow, on the other hand, is driven by pressure differentials and wall interactions, which makes the turbulence inherently anisotropic. In such cases, the relaxed group convolution is preferable, as it adeptly balances between upholding symmetry principles and adapting to asymmetric factors. All details are in Appendix \ref{app:superresolution}.

\section{Discussion}
We theoretically show and empirically demonstrate that relaxed group convolution adapts to the highest level of equivariance that is consistent with the data distribution and find that the relaxed weights post-training uncover symmetry-breaking factors. We employ various relaxed group convolution architectures to uncover symmetry-breaking factors in different physical systems, including the phase transition of crystal structure, the isotropy and homogeneity breaking in turbulence, and the time-reversal symmetry breaking in pendulum systems. Our method can also perform symmetry discovery as we can uncover hidden symmetries in data by creating relaxed group convolution layers of the largest possible group. Thus, we include experiment results regarding symmetry discovery baselines in Appendix \ref{app:baselines}.

One potential limitation of our method is requires an assumption of the largest possible group beforehand but general physics usually provides a clear indication of the largest possible relevant symmetry group. Another limitation is that our method may fail under spontaneous symmetry breaking when the data contain multiple possible outputs with equal probability. In this case, one would need a probabilistic model that generates a set of outputs, which is one of our future research directions. However, we still want to emphasize that our method is effective in most symmetry-breaking settings. Future work includes generalizing the proposed method to graph neural nets and investigating the benefits of relaxed weights in optimization.

\section*{Impact Statement}
Our research on modeling symmetry breaking in physical systems using relaxed group convolutions marks a significant advancement in machine learning with applications in physics and engineering. It offers insights into phenomena from particle interactions to cosmic structures, with implications for material science and environmental modeling. Ethically, the potential dual-use of such advanced modeling techniques necessitates careful consideration to ensure their beneficial application to society. We advocate for ethical vigilance and responsible use of this technology as it continues to evolve.

\section*{Acknowledgement}
Rui Wang and Tess Smidt were supported by DOE ICDI grant DE-SC0022215. Robin Walters was supported by NSF 2134178. Elyssa Hofgard was supported by the U.S. Department of Energy, Office of Science, Office of Advanced Scientific Computing Research, Department of Energy Computational Science Graduate Fellowship under Award Number DE-SC0024386. This research used resources of the National Energy Research Scientific Computing Center (NERSC) under Award Number ERCAP0028753, a Department of Energy Office of Science User Facility. This work is supported by the National Science Foundation under Cooperative Agreement PHY-2019786 (The NSF AI Institute for Artificial Intelligence and Fundamental Interactions, \url{http://iaifi.org/}).
\newpage
\clearpage
\bibliographystyle{icml2024}
\bibliography{example_paper}

\begin{thebibliography}{67}
\providecommand{\natexlab}[1]{#1}
\providecommand{\url}[1]{\texttt{#1}}
\expandafter\ifx\csname urlstyle\endcsname\relax
  \providecommand{\doi}[1]{doi: #1}\else
  \providecommand{\doi}{doi: \begingroup \urlstyle{rm}\Url}\fi

\bibitem[Dre(2008)]{Dresselhaus2008}
\emph{Character of a Representation}, pp.\  29--55.
\newblock Springer Berlin Heidelberg, Berlin, Heidelberg, 2008.
\newblock \doi{10.1007/978-3-540-32899-5_3}.
\newblock URL \url{https://doi.org/10.1007/978-3-540-32899-5_3}.

\bibitem[Anderson et~al.(2019)Anderson, Hy, and Kondor]{Anderson2019Cormorant}
Anderson, B., Hy, T.-S., and Kondor, R.
\newblock Cormorant: Covariant molecular neural networks.
\newblock In \emph{Advances in neural information processing systems (NeurIPS)}, 2019.

\bibitem[Bao \& Song(2019)Bao and Song]{bao2019equivariant}
Bao, E. and Song, L.
\newblock Equivariant neural networks and equivarification.
\newblock \emph{arXiv preprint arXiv:1906.07172}, 2019.

\bibitem[Batchelor(1953)]{batchelor1953theory}
Batchelor, G.~K.
\newblock \emph{The theory of homogeneous turbulence}.
\newblock Cambridge university press, 1953.

\bibitem[Beekman et~al.(2019)Beekman, Rademaker, and van Wezel]{beekman2019introduction}
Beekman, A., Rademaker, L., and van Wezel, J.
\newblock An introduction to spontaneous symmetry breaking.
\newblock \emph{SciPost Physics Lecture Notes}, pp.\  011, 2019.

\bibitem[Benidis et~al.(2022)Benidis, Rangapuram, Flunkert, Wang, Maddix, Turkmen, Gasthaus, Bohlke-Schneider, Salinas, Stella, et~al.]{benidis2022deep}
Benidis, K., Rangapuram, S.~S., Flunkert, V., Wang, Y., Maddix, D., Turkmen, C., Gasthaus, J., Bohlke-Schneider, M., Salinas, D., Stella, L., et~al.
\newblock Deep learning for time series forecasting: Tutorial and literature survey.
\newblock \emph{ACM Computing Surveys}, 55\penalty0 (6):\penalty0 1--36, 2022.

\bibitem[Bradley \& Cracknell(2010)Bradley and Cracknell]{bradley2010mathematical}
Bradley, C. and Cracknell, A.
\newblock \emph{The mathematical theory of symmetry in solids: representation theory for point groups and space groups}.
\newblock Oxford University Press, 2010.

\bibitem[Bronstein et~al.(2021)Bronstein, Bruna, Cohen, and Veli{\v{c}}kovi{\'c}]{bronstein2021geometric}
Bronstein, M.~M., Bruna, J., Cohen, T., and Veli{\v{c}}kovi{\'c}, P.
\newblock Geometric deep learning: Grids, groups, graphs, geodesics, and gauges.
\newblock \emph{arXiv:2104.13478}, 2021.

\bibitem[Castellani \& Dardashti(2021)Castellani and Dardashti]{Castellani2021symmbreak}
Castellani, E. and Dardashti, R.
\newblock Symmetry breaking.
\newblock In Knox, E. and Wilson, A. (eds.), \emph{The Routledge Companion to Philosophy of Physics}. Routledge, 2021.

\bibitem[Chidester et~al.(2018)Chidester, Do, and Ma]{chidester2018rotation}
Chidester, B., Do, M.~N., and Ma, J.
\newblock Rotation equivariance and invariance in convolutional neural networks.
\newblock \emph{arXiv preprint arXiv:1805.12301}, 2018.

\bibitem[Cohen \& Welling(2016{\natexlab{a}})Cohen and Welling]{Cohen2016Group}
Cohen, T. and Welling, M.
\newblock Group equivariant convolutional networks.
\newblock In \emph{Proceedings of the International Conference on Machine Learning (ICML)}, 2016{\natexlab{a}}.

\bibitem[Cohen \& Welling(2016{\natexlab{b}})Cohen and Welling]{cohen2016steerable}
Cohen, T.~S. and Welling, M.
\newblock Steerable {CNN}s.
\newblock \emph{arXiv preprint arXiv:1612.08498}, 2016{\natexlab{b}}.

\bibitem[Cohen et~al.(2019)Cohen, Weiler, Kicanaoglu, and Welling]{cohen2019gauge}
Cohen, T.~S., Weiler, M., Kicanaoglu, B., and Welling, M.
\newblock Gauge equivariant convolutional networks and the icosahedral {CNN}.
\newblock In \emph{Proceedings of the 36th International Conference on Machine Learning (ICML)}, volume~97, pp.\  1321--1330, 2019.

\bibitem[Dehmamy et~al.(2021)Dehmamy, Walters, Liu, Wang, and Yu]{dehmamy2021automatic}
Dehmamy, N., Walters, R., Liu, Y., Wang, D., and Yu, R.
\newblock Automatic symmetry discovery with lie algebra convolutional network.
\newblock \emph{Advances in Neural Information Processing Systems}, 34:\penalty0 2503--2515, 2021.

\bibitem[Desai et~al.(2022)Desai, Nachman, and Thaler]{desai2022symmetry}
Desai, K., Nachman, B., and Thaler, J.
\newblock Symmetry discovery with deep learning.
\newblock \emph{Physical Review D}, 105\penalty0 (9):\penalty0 096031, 2022.

\bibitem[Elsayed et~al.(2020)Elsayed, Ramachandran, Shlens, and Kornblith]{Elsayed2020Revisiting}
Elsayed, G.~F., Ramachandran, P., Shlens, J., and Kornblith, S.
\newblock Revisiting spatial invariance with low-rank local connectivity.
\newblock In \emph{Proceedings of the 37th International Conference of Machine Learning (ICML)}, 2020.

\bibitem[Esteves et~al.(2018)Esteves, Allen-Blanchette, Makadia, and Daniilidis]{esteves2018learning}
Esteves, C., Allen-Blanchette, C., Makadia, A., and Daniilidis, K.
\newblock Learning so (3) equivariant representations with spherical cnns.
\newblock In \emph{Proceedings of the European Conference on Computer Vision (ECCV)}, pp.\  52--68, 2018.

\bibitem[Fahle et~al.(2001)Fahle, Schamberger, and Sellmann]{fahle2001symmetry}
Fahle, T., Schamberger, S., and Sellmann, M.
\newblock Symmetry breaking.
\newblock In \emph{Principles and Practice of Constraint Programming—CP 2001: 7th International Conference, CP 2001 Paphos, Cyprus, November 26--December 1, 2001 Proceedings 7}, pp.\  93--107. Springer, 2001.

\bibitem[Finn et~al.(2017)Finn, Abbeel, and Levine]{finn2017model}
Finn, C., Abbeel, P., and Levine, S.
\newblock Model-agnostic meta-learning for fast adaptation of deep networks.
\newblock In \emph{International conference on machine learning}, pp.\  1126--1135. PMLR, 2017.

\bibitem[Finzi et~al.(2020)Finzi, Stanton, Izmailov, and Wilson]{finzi2020generalizing}
Finzi, M., Stanton, S., Izmailov, P., and Wilson, A.~G.
\newblock Generalizing convolutional neural networks for equivariance to lie groups on arbitrary continuous data.
\newblock \emph{arXiv preprint arXiv:2002.12880}, 2020.

\bibitem[Finzi et~al.(2021)Finzi, Benton, and Wilson]{finzi2021residual}
Finzi, M.~A., Benton, G., and Wilson, A.~G.
\newblock Residual pathway priors for soft equivariance constraints.
\newblock In Beygelzimer, A., Dauphin, Y., Liang, P., and Vaughan, J.~W. (eds.), \emph{Advances in Neural Information Processing Systems}, 2021.
\newblock URL \url{https://openreview.net/forum?id=k505ekjMzww}.

\bibitem[Gao et~al.(2023)Gao, Han, Fan, Sun, Liu, Duan, and Wang]{gao2023bayesian}
Gao, H., Han, X., Fan, X., Sun, L., Liu, L.-P., Duan, L., and Wang, J.-X.
\newblock Bayesian conditional diffusion models for versatile spatiotemporal turbulence generation.
\newblock \emph{arXiv preprint arXiv:2311.07896}, 2023.

\bibitem[Geiger \& Smidt(2022)Geiger and Smidt]{geiger2022e3nn}
Geiger, M. and Smidt, T.
\newblock e3nn: Euclidean neural networks.
\newblock \emph{arXiv preprint arXiv:2207.09453}, 2022.

\bibitem[Ghosh \& Gupta(2019)Ghosh and Gupta]{Ghosh19Scale}
Ghosh, R. and Gupta, A.~K.
\newblock Scale steerable filters for locally scale-invariant convolutional neural networks.
\newblock \emph{arXiv preprint arXiv:1906.03861}, 2019.

\bibitem[Helwig et~al.(2023)Helwig, Zhang, Fu, Kurtin, Wojtowytsch, and Ji]{helwig2023group}
Helwig, J., Zhang, X., Fu, C., Kurtin, J., Wojtowytsch, S., and Ji, S.
\newblock Group equivariant fourier neural operators for partial differential equations.
\newblock \emph{arXiv preprint arXiv:2306.05697}, 2023.

\bibitem[Holl et~al.(2020)Holl, Um, and Thuerey]{phiflow}
Holl, P.~M., Um, K., and Thuerey, N.
\newblock phiflow: A differentiable pde solving framework for deep learning via physical simulations.
\newblock In \emph{Workshop on Differentiable Vision, Graphics, and Physics in Machine Learning at NeurIPS}, 2020.

\bibitem[Jain et~al.(2013)Jain, Ong, Hautier, Chen, Richards, Dacek, Cholia, Gunter, Skinner, Ceder, et~al.]{jain2013commentary}
Jain, A., Ong, S.~P., Hautier, G., Chen, W., Richards, W.~D., Dacek, S., Cholia, S., Gunter, D., Skinner, D., Ceder, G., et~al.
\newblock Commentary: The materials project: A materials genome approach to accelerating materials innovation.
\newblock \emph{APL materials}, 1\penalty0 (1), 2013.

\bibitem[Jucha et~al.(2014)Jucha, Xu, Pumir, and Bodenschatz]{jucha2014time}
Jucha, J., Xu, H., Pumir, A., and Bodenschatz, E.
\newblock Time-reversal-symmetry breaking in turbulence.
\newblock \emph{Physical review letters}, 113\penalty0 (5):\penalty0 054501, 2014.

\bibitem[Kanov et~al.(2015)Kanov, Burns, Lalescu, and Eyink]{kanov2015johns}
Kanov, K., Burns, R., Lalescu, C., and Eyink, G.
\newblock The johns hopkins turbulence databases: An open simulation laboratory for turbulence research.
\newblock \emph{Computing in Science \& Engineering}, 17\penalty0 (5):\penalty0 10--17, 2015.

\bibitem[Knigge et~al.(2022)Knigge, Romero, and Bekkers]{knigge2022exploiting}
Knigge, D.~M., Romero, D.~W., and Bekkers, E.~J.
\newblock Exploiting redundancy: Separable group convolutional networks on lie groups.
\newblock In \emph{International Conference on Machine Learning}, pp.\  11359--11386. PMLR, 2022.

\bibitem[Kondor \& Trivedi(2018)Kondor and Trivedi]{kondor2018generalization}
Kondor, R. and Trivedi, S.
\newblock On the generalization of equivariance and convolution in neural networks to the action of compact groups.
\newblock In \emph{Proceedings of the 35th International Conference on Machine Learning (ICML)}, volume~80, pp.\  2747--2755, 2018.

\bibitem[Lamb \& Roberts(1998)Lamb and Roberts]{lamb1998time}
Lamb, J.~S. and Roberts, J.~A.
\newblock Time-reversal symmetry in dynamical systems: a survey.
\newblock \emph{Physica D: Nonlinear Phenomena}, 112\penalty0 (1-2):\penalty0 1--39, 1998.

\bibitem[Landau(1936)]{landau1936orderparam}
Landau, L.
\newblock The theory of phase transitions.
\newblock \emph{Nature}, 138\penalty0 (3498):\penalty0 840--841, 1936.

\bibitem[Liao et~al.(2023)Liao, Wood, Das, and Smidt]{liao2023equiformerv2}
Liao, Y.-L., Wood, B., Das, A., and Smidt, T.
\newblock Equiformerv2: Improved equivariant transformer for scaling to higher-degree representations.
\newblock \emph{arXiv preprint arXiv:2306.12059}, 2023.

\bibitem[Liu et~al.(2022)Liu, Wang, Liu, Lin, Zhang, Oztekin, and Ji]{liu2022spherical}
Liu, Y., Wang, L., Liu, M., Lin, Y., Zhang, X., Oztekin, B., and Ji, S.
\newblock Spherical message passing for 3d molecular graphs.
\newblock In \emph{International Conference on Learning Representations}, 2022.
\newblock URL \url{https://openreview.net/forum?id=givsRXsOt9r}.

\bibitem[Luke et~al.(1998)Luke, Fudamoto, Kojima, Larkin, Merrin, Nachumi, Uemura, Maeno, Mao, Mori, et~al.]{luke1998time}
Luke, G.~M., Fudamoto, Y., Kojima, K., Larkin, M., Merrin, J., Nachumi, B., Uemura, Y., Maeno, Y., Mao, Z., Mori, Y., et~al.
\newblock Time-reversal symmetry-breaking superconductivity in sr2ruo4.
\newblock \emph{Nature}, 394\penalty0 (6693):\penalty0 558--561, 1998.

\bibitem[McNeela(2023)]{mcneela2023almost}
McNeela, D.
\newblock Almost equivariance via lie algebra convolutions.
\newblock \emph{arXiv preprint arXiv:2310.13164}, 2023.

\bibitem[Nabarro(1947)]{nabarro1947dislocations}
Nabarro, F.
\newblock Dislocations in a simple cubic lattice.
\newblock \emph{Proceedings of the Physical Society}, 59\penalty0 (2):\penalty0 256, 1947.

\bibitem[Onuki(2002)]{onuki2002phase}
Onuki, A.
\newblock \emph{Phase transition dynamics}.
\newblock Cambridge University Press, 2002.

\bibitem[Otto et~al.(2023)Otto, Zolman, Kutz, and Brunton]{otto2023unified}
Otto, S.~E., Zolman, N., Kutz, J.~N., and Brunton, S.~L.
\newblock A unified framework to enforce, discover, and promote symmetry in machine learning.
\newblock \emph{arXiv preprint arXiv:2311.00212}, 2023.

\bibitem[Passaro \& Zitnick(2023)Passaro and Zitnick]{passaro2023reducing}
Passaro, S. and Zitnick, C.~L.
\newblock Reducing so (3) convolutions to so (2) for efficient equivariant gnns.
\newblock \emph{arXiv preprint arXiv:2302.03655}, 2023.

\bibitem[Petrache \& Trivedi(2023)Petrache and Trivedi]{petrache2023approximation}
Petrache, M. and Trivedi, S.
\newblock Approximation-generalization trade-offs under (approximate) group equivariance.
\newblock \emph{arXiv preprint arXiv:2305.17592}, 2023.

\bibitem[Pope(2001)]{pope2001turbulent}
Pope, S.~B.
\newblock Turbulent flows.
\newblock \emph{Measurement Science and Technology}, 12\penalty0 (11):\penalty0 2020--2021, 2001.

\bibitem[Satorras et~al.(2021{\natexlab{a}})Satorras, Hoogeboom, and Welling]{Garcia2021EN}
Satorras, V.~G., Hoogeboom, E., and Welling, M.
\newblock E(n) equivariant graph neural networks.
\newblock \emph{arXiv preprint arXiv:2102.09844}, 2021{\natexlab{a}}.

\bibitem[Satorras et~al.(2021{\natexlab{b}})Satorras, Hoogeboom, and Welling]{satorras2021n}
Satorras, V.~G., Hoogeboom, E., and Welling, M.
\newblock E (n) equivariant graph neural networks.
\newblock In \emph{International Conference on Machine Learning}, pp.\  9323--9332. PMLR, 2021{\natexlab{b}}.

\bibitem[Sch{\"u}tt et~al.(2017)Sch{\"u}tt, Kindermans, Sauceda~Felix, Chmiela, Tkatchenko, and M{\"u}ller]{schutt2017schnet}
Sch{\"u}tt, K., Kindermans, P.-J., Sauceda~Felix, H.~E., Chmiela, S., Tkatchenko, A., and M{\"u}ller, K.-R.
\newblock Schnet: A continuous-filter convolutional neural network for modeling quantum interactions.
\newblock \emph{Advances in neural information processing systems}, 30, 2017.

\bibitem[Smidt et~al.(2021)Smidt, Geiger, and Miller]{smidt2021finding}
Smidt, T.~E., Geiger, M., and Miller, B.~K.
\newblock Finding symmetry breaking order parameters with euclidean neural networks.
\newblock \emph{Physical Review Research}, 3\penalty0 (1):\penalty0 L012002, 2021.

\bibitem[Strocchi(2005)]{strocchi2005symmetry}
Strocchi, F.
\newblock \emph{Symmetry breaking}, volume 643.
\newblock Springer, 2005.

\bibitem[Thomas et~al.(2018)Thomas, Smidt, Kearnes, Yang, Li, Kohlhoff, and Riley]{thomas2018tensor}
Thomas, N., Smidt, T., Kearnes, S., Yang, L., Li, L., Kohlhoff, K., and Riley, P.
\newblock Tensor field networks: Rotation-and translation-equivariant neural networks for 3d point clouds.
\newblock \emph{arXiv preprint arXiv:1802.08219}, 2018.

\bibitem[Van~der Laan \& Kirkman(1992)Van~der Laan and Kirkman]{van19922p}
Van~der Laan, G. and Kirkman, I.
\newblock The 2p absorption spectra of 3d transition metal compounds in tetrahedral and octahedral symmetry.
\newblock \emph{Journal of Physics: Condensed Matter}, 4\penalty0 (16):\penalty0 4189, 1992.

\bibitem[van~der Ouderaa et~al.(2023)van~der Ouderaa, Immer, and van~der Wilk]{van2023learning}
van~der Ouderaa, T.~F., Immer, A., and van~der Wilk, M.
\newblock Learning layer-wise equivariances automatically using gradients.
\newblock \emph{arXiv preprint arXiv:2310.06131}, 2023.

\bibitem[Walters et~al.(2021)Walters, Li, and Yu]{Walters2021ECCO}
Walters, R., Li, J., and Yu, R.
\newblock Trajectory prediction using equivariant continuous convolution.
\newblock \emph{International Conference on Learning Representations}, 2021.

\bibitem[Wang et~al.(2020)Wang, Walters, and Yu]{wang2020incorporating}
Wang, R., Walters, R., and Yu, R.
\newblock Incorporating symmetry into deep dynamics models for improved generalization.
\newblock In \emph{International Conference on Learning Representations (ICLR)}, 2020.

\bibitem[Wang et~al.(2022)Wang, Walters, and Yu]{wang2022approximately}
Wang, R., Walters, R., and Yu, R.
\newblock Approximately equivariant networks for imperfectly symmetric dynamics.
\newblock In \emph{International Conference on Machine Learning}, pp.\  23078--23091. PMLR, 2022.

\bibitem[Weiler \& Cesa(2019)Weiler and Cesa]{weiler2019e2cnn}
Weiler, M. and Cesa, G.
\newblock General {E}(2)-equivariant steerable {CNNs}.
\newblock In \emph{Advances in Neural Information Processing Systems (NeurIPS)}, pp.\  14334--14345, 2019.

\bibitem[Weinberg(1976)]{weinberg1976implications}
Weinberg, S.
\newblock Implications of dynamical symmetry breaking.
\newblock \emph{Physical Review D}, 13\penalty0 (4):\penalty0 974, 1976.

\bibitem[Woodward(1997)]{woodward1997octahedral}
Woodward, P.~M.
\newblock Octahedral tilting in perovskites. i. geometrical considerations.
\newblock \emph{Acta Crystallographica Section B: Structural Science}, 53\penalty0 (1):\penalty0 32--43, 1997.

\bibitem[Worrall et~al.(2017)Worrall, Garbin, Turmukhambetov, and Brostow]{worrall2017harmonic}
Worrall, D.~E., Garbin, S.~J., Turmukhambetov, D., and Brostow, G.~J.
\newblock Harmonic networks: Deep translation and rotation equivariance.
\newblock In \emph{Proceedings of the IEEE Conference on Computer Vision and Pattern Recognition}, pp.\  5028--5037, 2017.

\bibitem[Yang et~al.(2023{\natexlab{a}})Yang, Dehmamy, Walters, and Yu]{yang2023latent}
Yang, J., Dehmamy, N., Walters, R., and Yu, R.
\newblock Latent space symmetry discovery.
\newblock \emph{arXiv preprint arXiv:2310.00105}, 2023{\natexlab{a}}.

\bibitem[Yang et~al.(2023{\natexlab{b}})Yang, Walters, Dehmamy, and Yu]{yang2023generative}
Yang, J., Walters, R., Dehmamy, N., and Yu, R.
\newblock Generative adversarial symmetry discovery.
\newblock \emph{arXiv preprint arXiv:2302.00236}, 2023{\natexlab{b}}.

\bibitem[Ying et~al.(2018)Ying, Yuan, Vlaski, and Sayed]{ying2018stochastic}
Ying, B., Yuan, K., Vlaski, S., and Sayed, A.~H.
\newblock Stochastic learning under random reshuffling with constant step-sizes.
\newblock \emph{IEEE Transactions on Signal Processing}, 67\penalty0 (2):\penalty0 474--489, 2018.

\bibitem[Zaheer et~al.(2017)Zaheer, Kottur, Ravanbakhsh, Poczos, Salakhutdinov, and Smola]{zaheer2017deep}
Zaheer, M., Kottur, S., Ravanbakhsh, S., Poczos, B., Salakhutdinov, R., and Smola, A.
\newblock Deep sets.
\newblock \emph{arXiv preprint arXiv:1703.06114}, 2017.

\bibitem[Zakharov et~al.(2012)Zakharov, L'vov, and Falkovich]{zakharov2012kolmogorov}
Zakharov, V.~E., L'vov, V.~S., and Falkovich, G.
\newblock \emph{Kolmogorov spectra of turbulence I: Wave turbulence}.
\newblock Springer Science \& Business Media, 2012.

\bibitem[Zee(2016)]{zee2016group}
Zee, A.
\newblock \emph{Group theory in a nutshell for physicists}.
\newblock In a nutshell. Princeton University Press, 2016.
\newblock ISBN 978-0-691-16269-0.
\newblock URL \url{https://books.google.com/books?id=FWkujgEACAAJ}.
\newblock tex.lccn: 2015037408.

\bibitem[Zhang(1988)]{zhang1988shift}
Zhang, W.
\newblock Shift-invariant pattern recognition neural network and its optical architecture.
\newblock In \emph{Proceedings of annual conference of the Japan Society of Applied Physics}, 1988.

\bibitem[Zhou et~al.(2021)Zhou, Knowles, and Finn]{zhou2021metalearning}
Zhou, A., Knowles, T., and Finn, C.
\newblock Meta-learning symmetries by reparameterization.
\newblock In \emph{International Conference on Learning Representations}, 2021.
\newblock URL \url{https://openreview.net/forum?id=-QxT4mJdijq}.

\bibitem[Zitnick et~al.(2022)Zitnick, Das, Kolluru, Lan, Shuaibi, Sriram, Ulissi, and Wood]{zitnick2022spherical}
Zitnick, L., Das, A., Kolluru, A., Lan, J., Shuaibi, M., Sriram, A., Ulissi, Z., and Wood, B.
\newblock Spherical channels for modeling atomic interactions.
\newblock \emph{Advances in Neural Information Processing Systems}, 35:\penalty0 8054--8067, 2022.

\end{thebibliography}

\newpage
\appendix
\onecolumn
\section*{Appendix}
\section{Theoretical Analysis}\label{app:theory}
\begin{proposition}
Consider a relaxed group convolutional neural network $f$ where the relaxed weights in each layer are initialized to be identical across group elements to maintain $G$-equivariance. If $f$ is trained to map an input $x$ to output $y$, the relaxed weights will change during training such that $f$ is equivariant to $\text{Stab}(x) \cap \text{Stab}(y)$, which is the intersection of the stabilizers of the input and the output.   
\end{proposition}
\begin{proof}
Let $G$ be the semi-direct product of the translation group $(\mathbb{Z}^3, +)$ and a group $H$. Without loss of the generality, we only consider the composition of one lift convolution layer and a relaxed group convolution layer with a single filter bank case (i.e. $L=1$). Suppose the input is $f_0(\bm{x})$ and $\phi$ is an unconstrained kernel, then the output of the lift convolution layer is:

$$f_1(\bm{y}, h) = \sum_{\bm{x}\in\mathbb{Z}^3}f_0(\bm{x}) \phi(h^{-1}(\bm{x}-\bm{y})), \;\; h \in G$$

Now we prove that $f_1(\bm{y}, h) = f_1(k\bm{y}, kh),\;k \in G$ only when $k$ stabilizes $f_0$. 
\begin{equation}
\begin{aligned}
f_1(k\bm{y}, kh) &=  \sum_{\bm{x}\in\mathbb{Z}^3}f_0(\bm{x}) \phi((kh)^{-1}(\bm{x}-k\bm{y}))\\
&=  \sum_{\bm{x}\in\mathbb{Z}^3}f_0(\bm{x}) \phi(h^{-1}(k^{-1}\bm{x}-\bm{y})) \\
&=  \sum_{k\bm{x}\in\mathbb{Z}^3}f_0(k\bm{x}) \phi(h^{-1}(\bm{x}-\bm{y})) \\
\end{aligned}\nonumber
\end{equation}
Thus,  $f_1(\bm{y}, h) = f_1(k\bm{y}, kh)$ only when $f_0(k\bm{x}) = f_0(\bm{x})$.

We denote $f_2(\bm{z}, k)$ as the output of the group convolution layer and $ \psi$ as the kernel.
$$f_2(\bm{z}, k) = \sum_{\bm{y}\in \mathbb{Z}^3}\sum_{h\in G}f_1(\bm{y}, h) \psi(k^{-1}(\bm{y}-\bm{z}), k^{-1}h)$$

Now we prove that, $f_2(g\bm{z}, g) = f_2(\bm{z}, e),\;g \in G$ only when $g$ stabilizes $f_0$, where $e$ is the identity. 
\begin{equation}
\begin{aligned}
f_2(g\bm{z}, g) & = \sum_{\bm{y}\in \mathbb{Z}^3}\sum_{h\in G}f_1(\bm{y}, h) \psi(g^{-1}(\bm{y}-g\bm{z}), g^{-1}h) \\
 & = \sum_{g\bm{y}\in \mathbb{Z}^3}\sum_{gh\in G}f_1(g\bm{y}, gh) \psi(\bm{y}-\bm{z}), h)
\end{aligned}\nonumber
\end{equation}
Thus, $f_2(g\bm{z}, g) = f_2(\bm{z}, e)$ only when $f_1(g\bm{y}, gh) = f_1(\bm{y}, h)$, which means $f_0$ needs to be stablized by $g$  given the previous step of the proof. 

Let $Y(\bm{z})$ and $\hat{Y}(\bm{z})$, $\bm{z} \in \mathbb{Z}^3$,  be the target and prediction respectively and $L$ is MSE loss. In the last layer, we usually average over the $H$-axis and we can define $\hat{Y}(\bm{x})$ based on the definition of relaxed group convolution as follows:

$$\hat{Y}(\bm{z}) = \sum_{k\in H}w(k)f_2(\bm{z}, k)$$

where $w(k)$ are the relaxed weights. We use MSE loss:
$$L = \sum_{\bm{z}\in \mathbb{Z}^3} (Y(\bm{z}) - \hat{Y}(\bm{z}))^2$$

Finally, we can compute the gradient of the loss $L$ w.r.t a relaxed weight $w(k)$:
\begin{equation}
\begin{aligned}
\frac{\partial L}{\partial w(k)} &= \sum_{\bm{z}\in \mathbb{Z}^3}\frac{\partial L}{\partial \hat{Y}(\bm{z})}\frac{\partial \hat{Y}(\bm{z})}{\partial w(k)} \\
& = -2\sum_{\bm{z}\in \mathbb{Z}^3}|Y(\bm{z}) - \sum_{t\in H}  f_2(\bm{z}, t)| f_2(\bm{z}, k) \\
& = -2\sum_{k\bm{z}\in \mathbb{Z}^3}|Y(k\bm{z}) - \sum_{t\in H}  f_2(k\bm{z}, t)| f_2(k\bm{z}, k)
\end{aligned}\nonumber
\end{equation}

This means if $k$ does not stabilize the input $f_0$, then $f_2(\bm{kz}, k) \neq f_2(\bm{z}, e)$, i.e. $\frac{\partial L}{\partial w(k)} \neq \frac{\partial L}{\partial w(e)}$

If $k$ stabilizes the input $f_0$, then 
\begin{equation}
\begin{aligned}
\frac{\partial L}{\partial w(k)}
& = -2\sum_{k\bm{z}\in \mathbb{Z}^3}(Y(k\bm{z}) - \sum_{t\in H}  f_2(\bm{z}, k^{-1}t)) f_2(\bm{z}, e) \\
& = -2\sum_{\bm{z}\in \mathbb{Z}^3}(Y(k\bm{z}) - \sum_{t\in H}  f_2(\bm{z}, t)) f_2(\bm{z}, e) \\
\end{aligned}\nonumber
\end{equation}
Then we can see $\frac{\partial L}{\partial w(k)} = \frac{\partial L}{\partial w(e)}$ only when $Y(k\bm{z}) = Y(\bm{z})$. 

In conclusion, $\frac{\partial L}{\partial w(k)} = \frac{\partial L}{\partial w(e)}$ only when $k$ stabilizes both the input $f_0$ and the target $Y$. 
\end{proof}


\begin{proposition}\label{app:minibatch}
Denote $\mu$ as the step size of the gradient 
descent, $M$ as the number of mini-batches in each epoch, and $t$ as the number of epochs respectively. Assume that the loss 
function $L$ is convex and has $\delta$-Liptischitz continuous gradients. Consider $\bm{w}^*$ as the optimal relaxed weights that can be reached by full batch gradient descent, with $K$ representing the gradient noise variance at $\bm{w}^*$. Then the starting relaxed weights $\bm{w}^t_0$ at the $t$-th epoch satisfies
$$E\|\bm{w}^t_0 - \bm{w}^*\| \leq (\frac{\mu^2\delta^2 M^2-1}{2\mu^2\delta^2 M^2-1})^tE\|\bm{w}^0_0 - \bm{w}^*\| + \frac{\mu^4\delta^2 M^4 K}{1-2\mu^2\delta^2 M^2}$$
\end{proposition}
\begin{proof}
Denote $\mu$ as the step size of the gradient 
descent, $M$ as the number of mini-batches in each epoch, , $t$ as the number of epochs, $B$ as the batch size, and $N$ as the total number of training samples. Let $z_i = \{(x_j, y_j)\}_{j=B*i}^{B*(i+1)}$ be a mini-batch of training samples. 
The optimal relaxed weights $\bm{w}^*$ can be defined as 
$$\bm{w}^* = \text{argmin} J(\bm{w}) = \text{argmin} \frac{1}{M}\sum_{i=1}^{M} L(\bm{w}; z_i)$$

\textit{Assumption 1}: we assume that $L(\bm{w}; z)$ is differentiable and has $\delta$-Liptischitz continous gradients, i.e.
$$\|\nabla_{\bm{w}}L(\bm{w}_1; z_i)-\nabla_{\bm{w}}L(\bm{w}_2; z_i)\| \leq \delta\|\bm{w}_1-\bm{w}_2\|, \;\;  i = 1, ..., M$$

\textit{Assumption 2}: we assume $J(\bm{w})$ is convex, for any $\bm{w}_1, \bm{w}_2$:
$$(J_{\bm{w}_1}(\bm{w}_1)-J_{\bm{w}_2}(\bm{w}_2))^T(\bm{w}_1 - \bm{w}_2) \geq 0$$

We denote $\bm{w}_i^t$ as the relaxed weights at iteration $i$ in $t$-th epoch. That means, 
$$\bm{w}_i^t = \bm{w}_{i-1}^t - \mu \nabla_{\bm{w}}L(\bm{w}_{i-1}^t, z_i^t)$$

\begin{align*}
  \bm{w}_0^{t+1} = \bm{w}_M^{t} & = \bm{w}_{M-1}^{t} - \mu \nabla_{\bm{w}}L(\bm{w}_{M-1}^t, z_M^t) \\
  & = \bm{w}_{0}^{t} - \mu \sum_{i=1}^M\nabla_{\bm{w}}L(\bm{w}_{i-1}^t, z_i^t) \\
  & = \bm{w}_{0}^{t} - \mu M\nabla_{\bm{w}}J(\bm{w}_0^k)-\mu \sum_{i=1}^M[\nabla_{\bm{w}}L(\bm{w}_{i-1}^t, z_i^t)-\nabla_{\bm{w}}L(\bm{w}_0^t, z_i^t)]
\end{align*}
We denote the last term $\sum_{i=1}^M[\nabla_{\bm{w}}L(\bm{w}_{i-1}^t, z_i^t)-\nabla_{\bm{w}}L(\bm{w}_0^t, z_i^t)]$ as $\epsilon_i^t(\bm{w}_{i-1}^t)$, which is the gradient mismatch between using mini-batch gradient descent and full batch gradient descent. 

Let $\tilde{\bm{w}}_{0}^t = \bm{w}^* - \bm{w}^t_0$ and subtract $\bm{w}^*$ from both sides,
\begin{align*}
\|\tilde{\bm{w}}_0^{t+1}\|^2 &= \|\tilde{\bm{w}}_0^{t} + \mu M \nabla_{\bm{w}} J(\bm{w}_0^t) + \mu \sum_{i=1}^M \epsilon_i^t(\bm{w}_{i-1}^t)\|^2 \\
& \leq \|\tilde{\bm{w}}_0^{t} + \mu M \nabla_{\bm{w}} J(\bm{w}_0^t) \|^2 + \|\mu \sum_{i=1}^M \epsilon_i^t(\bm{w}_{i-1}^t)\|^2 \\
& \leq \|\tilde{\bm{w}}_0^{t} + \mu M \nabla_{\bm{w}} J(\bm{w}_0^t) \|^2 + \mu^2 M \sum_{i=1}^M\|\epsilon_i^t(\bm{w}_{i-1}^t)\|^2 \\
\end{align*}

For the first term, we have 
\begin{align*}
\|\tilde{\bm{w}}_0^{t} + \mu M \nabla_{\bm{w}} J(\bm{w}_0^t) \|^2 & = \|\tilde{\bm{w}}_0^{t}\|^2 + \mu^2 M^2 \|\nabla_{\bm{w}} J(\bm{w}_0^t) \|^2 + 2\mu M (\tilde{\bm{w}}_0^{t})^T \nabla_{\bm{w}} J(\bm{w}_0^t) 
\end{align*}
Because of the convexity of $J$, we have 
$$
(\tilde{\bm{w}}_0^{t})^T \nabla_{\bm{w}} J(\bm{w}_0^t)  = (\bm{w}^* - \bm{w}_0^{t})^T (\nabla_{\bm{w}} J(\bm{w}_0^t) - \nabla_{\bm{w}} J(\bm{w}^*)) = -(\bm{w}_0^{t} - \bm{w}^*)^T (\nabla_{\bm{w}} J(\bm{w}_0^t) - \nabla_{\bm{w}} J(\bm{w}^*)) \leq 0
$$
Because $L$ has $\delta$-Liptischitz continuous gradients, we have 
$$\|\nabla_{\bm{w}} J(\bm{w}_0^t) \|^2 = \|\nabla_{\bm{w}} J(\bm{w}_0^t) - \nabla_{\bm{w}} J(\bm{w}^*) \|^2 \leq \delta\|\bm{w}_0^t - \bm{w}^*\|^2 = \delta^2\|\tilde{\bm{w}}_0^t\|^2 $$

For the second term $\sum_{i=1}^M\|\epsilon_i^t(\bm{w}_{i-1}^t)\|^2$, we can directly use the result in the Appendix B in \citet{ying2018stochastic},
$$\sum_{i=1}^M\|\epsilon_i^t(\bm{w}_{i-1}^t)\|^2 \leq \frac{\mu^2\delta^2 M^3}{1-\mu^2\delta^2 M^2}(2 \delta^2 \|\tilde{\bm{w}}_0^t\|^2 + K)$$
where $K = \frac{1}{M}\sum_{i=1}^M \|\nabla_{\bm{w}} L(\bm{w}^*, z_i) \|^2$. It is the gradient noise variance at optimal point $\bm{w}^*$. 

Thus, 
$$\|\tilde{\bm{w}}_0^{t+1}\|^2 \leq \frac{\mu^2\delta^2 M^2-1}{2\mu^2\delta^2 M^2-1} \|\tilde{\bm{w}}^t_0\|^2 + \frac{\mu^4\delta^2 M^4 K}{1-2\mu^2\delta^2 M^2}$$

Iterating over $t$, we have
\begin{align*}
\|\tilde{\bm{w}}_0^{t+1}\|^2 &\leq (\frac{\mu^2\delta^2 M^2-1}{2\mu^2\delta^2 M^2-1})^t \|\tilde{\bm{w}}^0_0\|^2 + (\frac{\mu^4\delta^2 M^4 K}{1-2\mu^2\delta^2 M^2})\sum_{j=1}^t(\frac{\mu^2\delta^2 M^2-1}{2\mu^2\delta^2 M^2-1})^j \\
& \leq  (\frac{\mu^2\delta^2 M^2-1}{2\mu^2\delta^2 M^2-1})^t \|\tilde{\bm{w}}^0_0\|^2 + (\frac{\mu^4\delta^2 M^4 K}{1-2\mu^2\delta^2 M^2})\sum_{j=1}^t(\frac{1}{2})^j \\
& \leq  (\frac{\mu^2\delta^2 M^2-1}{2\mu^2\delta^2 M^2-1})^t \|\tilde{\bm{w}}^0_0\|^2 + (\frac{\mu^4\delta^2 M^4 K}{1-2\mu^2\delta^2 M^2}) \\
& \leq  (1 - \frac{\mu^2\delta^2 M^2}{2\mu^2\delta^2 M^2-1})^t \|\tilde{\bm{w}}^0_0\|^2 + (\frac{\mu^4\delta^2 M^4 K}{1-2\mu^2\delta^2 M^2})
\end{align*}
\end{proof}

\section{Super-resolution of Velocity Fields in Three-dimensional Fluid Dynamics}\label{app:superresolution}
\paragraph{Data Description.}
We use the direct numerical simulation data of the channel flow ($2048\times512\times1536$) turbulence and the forced isotropic turbulence ($1024^3$) from Johns Hopkins Turbulence Database \citep{kanov2015johns}. For each dataset, we acquire 50 frames of velocity fields, which are then downscaled by half and segmented into $64^3$ cubes for experimental use. These cubes are further downsampled by a factor of 4 to serve as input for our superresolution model. The models are trained to generate $64^3$ simulations from $16^3$ downsampled versions of themselves. Because of the spatial weight sharing of CNNs, we can apply our model to 3D input with any resolution during inference. 
\begin{figure}[htb!]
	\centering
	\includegraphics[width=0.7\textwidth]{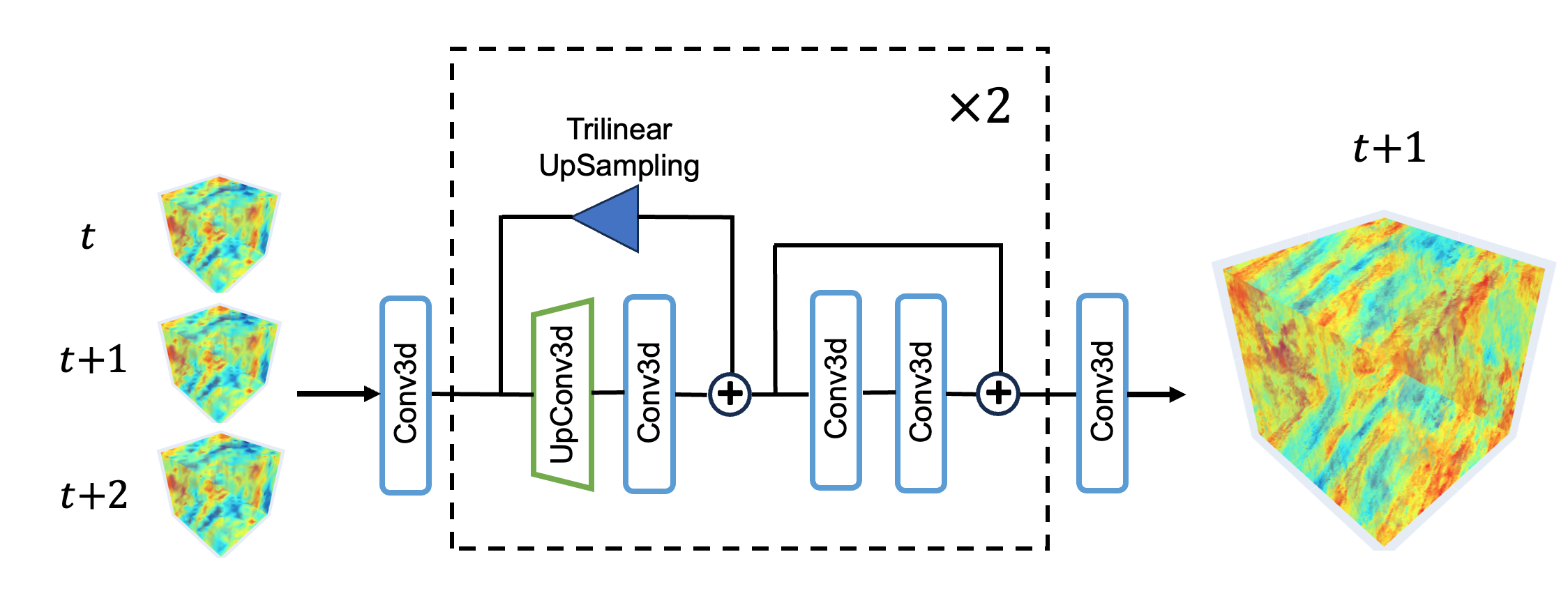}
	\caption{The architecture of the super-resolution model includes an input layer, an output layer, and eight residual blocks. Since it has two layers of Transposed Convolution(UpConv3d), the model produces simulations that are upscaled by a factor of four.}
	\label{fig:model}
\end{figure}

\begin{figure}[htb!]
	\centering
	\includegraphics[width=0.49\textwidth]{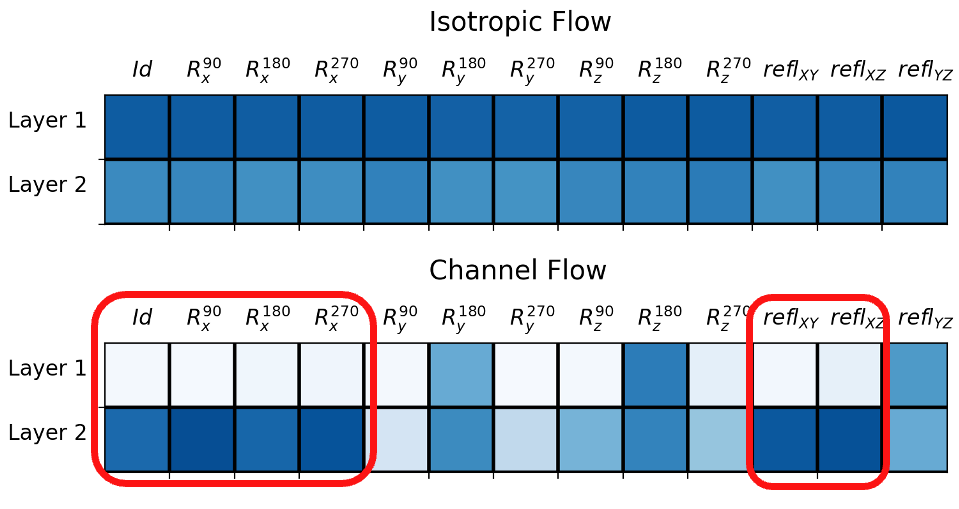}
	\caption{Visualization of the relaxed weights of first two layers from the models trained on the isotropic flow and channel flow.}
	\label{fig:vis_relaxedweights_flow}
\end{figure}

\newcolumntype{P}[1]{>{\centering\arraybackslash}p{#1}}
\begin{table*}[thb!]
\small
    \centering
    \caption{Prediction MAE of trilinear upsampling, non-equivariant, equivariant, relaxed equivariant models on the super-resolution of channel flow and isotropic flow.}
    \label{tab:superresoluion}
    \begin{tabular}{P{0.8cm}|P{1.3cm}P{1.3cm}P{1.3cm}P{1.7cm}|P{1.3cm}P{1.3cm}P{1.3cm}P{1.7cm}}\toprule
    & \multicolumn{4}{c|}{Channel Flow $(10^{-2})$} & \multicolumn{4}{c}{Isotropic Flow $(10^{-1})$}\\ \midrule
       Model  &  $\texttt{Trilinear}$ &  $\texttt{Conv}$ &  $\texttt{Equiv}$ & $\texttt{R-Equiv}$ & $\texttt{Trilinear}$ &  $\texttt{Conv}$ &  $\texttt{Equiv}$ & $\texttt{R-Equiv}$ \\ \midrule
       MAE  & 5.24 & 2.60$\pm$0.05 &  2.54$\pm$0.03  & $\bm{2.44\pm0.01}$ & 5.25 &  1.22$\pm$0.04 & 1.12$\pm$0.02  & $\bm{1.00\pm0.01}$ \\
        \bottomrule
    \end{tabular}
\end{table*}

\paragraph{Experimental Setup}
Figure \ref{fig:model} visualizes the model architecture we use for super-resolution. We evaluate the performance of Regular, Group Equivariant, and Relaxed Group Equivariant layers built into this architecture in the tasks of upscaling channel flow and isotropic turbulence. The models take three consecutive steps of downsampled $16^3$ velocity fields as input and predict a single step of $64^3$ simulation, enabling them to infer vital attributes like acceleration and external forces for precise small-scale turbulence predictions.  We use the L1 loss function over the L2 loss, as it significantly enhances performance. We split the data 80\%-10\%-10\% for training-validation-test across time and report mean absolute errors over three random runs. As for hyperparameter tuning, except for fixing the number of layers and kernel sizes, we perform a grid search for the learning rate, hidden dimensions, batch size, and the number of filter bases for all three types of models.

\begin{figure*}[htb!]
	\centering
	\includegraphics[width=\textwidth]{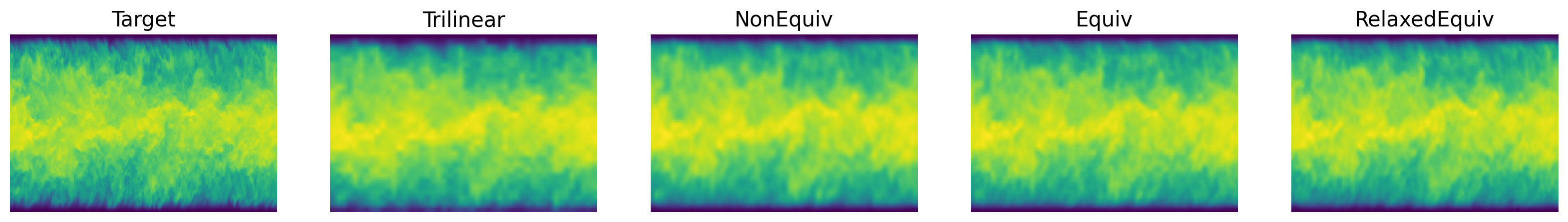}
 \includegraphics[width=\textwidth]{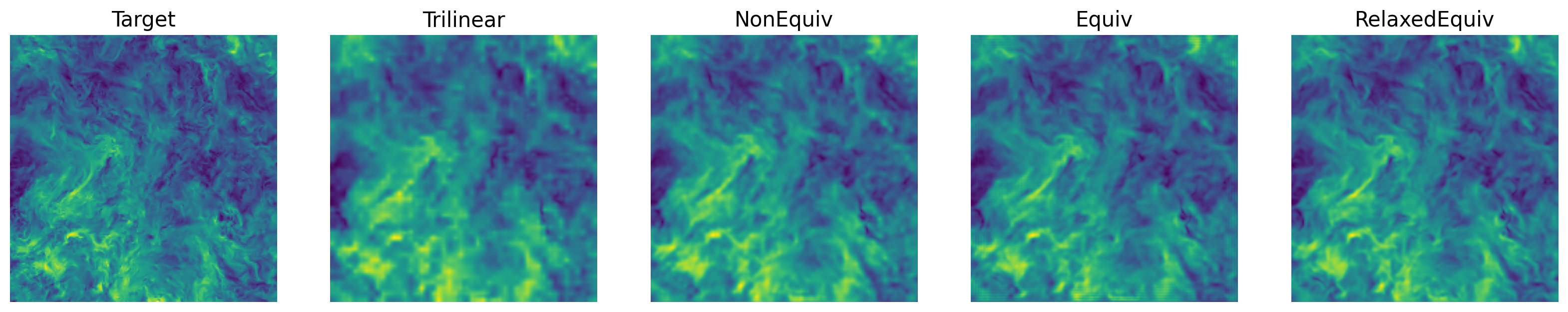}
	\caption{Prediction visualization of a cross-section along the z-axis of the velocity norm fields.}
	\label{fig:vis_flow}
\end{figure*}

\paragraph{Prediction Performance}
Table \ref{tab:superresoluion} shows the prediction MAE of trilinear upsampling, non-equivariant, equivariant, and relaxed equivariant models applied to super-resolution tasks for both channel and isotropic flows. Figure \ref{fig:vis_flow} shows the 2D velocity norm field of predictions. As we can see, imposing equivariance and relaxed equivariance consistently yields better prediction performance. It is fascinating to see that relaxed group convolution performs best even on the isotropic flow, which exhibits distributional rotation symmetry. We conjecture that relaxed weights may also enhance optimization, which we will leave for future work. 

Figure \ref{fig:vis_relaxedweights_flow} visualizes the relaxed weights of the first two layers from the models trained on isotropic flow and channel flow. The relaxed weights for isotropic flow stay almost the same across group elements while those for channel flow vary greatly, which conforms to the symmetries of these two types of flow. For isotropic turbulence, even if individual samples might not seem symmetrical, the statistical properties of their velocity fields over time and space are invariant with respect to rotations. This makes models trained on isotropic flow benefit more from the equivariance.

Channel flow, on the other hand, is driven by a pressure difference between the two ends of the channel together with the walls, which makes the turbulence inherently anisotropic. In such cases, the relaxed group convolution is preferable, as it adeptly balances between upholding certain symmetry principles and adapting to factors that introduce asymmetry. As we can see from the highlighted relaxed weights from Figure \ref{fig:vis_relaxedweights_flow}, though most symmetries are broken, the model can still learn approximate $D_4$ symmetry along the direction of the channel flow. 

Though group convolution has been a powerful tool for many applications, its computational complexity scales exponentially with the dimensionality of the group, which is the main practical challenge when dealing with the octahedral group. To improve parameter efficiency, we employ separable group convolution that assumes the convolution kernels are separable \citep{knigge2022exploiting}. 

\section{Irrep Analysis} \label{app:irrep_analysis}
\begin{proposition}
Consider a relaxed group convolution neural network for some group $G$ and $w \in \mathbb{R}^G$ is the relaxed weights from one of the layers. $w$ transforms under the regular representation $L_g \in \mathbb{R}^{G\times G}$ of the group $G$. $\{\rho_{\lambda}\}$ are the set of unique irreducible representations of $G$.  The Fourier transform of $w$ at irreducible representation $\rho_{\lambda}$ is defined as: $\hat{w}(\lambda) = \sum_g \rho_{\lambda}(g)w(g)$. We also define the stabilizer of $w$ as $\text{Stab}_G(w) = \{g | L_gw = w\}$ and the stabilizer of $\hat{w}(\lambda)$ as  $\text{Stab}_G(\hat{w}(\lambda)) = \{g | \rho_{\lambda}(g)\hat{w}(\lambda) = \hat{w}(\lambda)\}$. Then the stabilizer of $w$ is the same as the intersection of the stabilizers of the Fourier transforms across all irreducible representations, i.e. $\text{Stab}_G(w) = \cap_{\lambda}\text{Stab}_G(\hat{w}(\lambda))$.

Proof: The Fourier transform of $w$ on Group $G$, $\mathcal{F}: \mathbb{R}^G \rightarrow \bigoplus_{\lambda}\text{End}(V_{\lambda})$, is an isomorphism implied in the Wedderburn theorem and the inverse formula can be defined as:
$$w(g) = \frac{1}{|G|}\sum_{\lambda} d_{\lambda}tr(\hat{w}(\lambda)(\rho_{\lambda}(g))^{-1}) $$
where the $V_{\lambda}$ is subspace that $\rho_{\lambda}$ acts on and  $d_{\lambda}$ is the dimension of $V_{\lambda}$.

$\mathcal{F}$ is equivariant as well because 
$$ \rho_{\lambda}(h)\hat{w}(\lambda) = \sum_g \rho_{\lambda}(h)\rho_{\lambda}(g)w(g) = \sum_g \rho_{\lambda}(hg)w(g) = \sum_g \rho_{\lambda}(g)w(h^{-1}g) = \widehat{L_h w}(\lambda), \;\;\; \forall \lambda \in G $$

Thus, $\mathcal{F}(L_h w) = \bigoplus_{\lambda} \bigoplus_{d_{\lambda}} \rho_{\lambda}(h) vec(\hat{w}(\lambda)), \;\; \forall h \in G$. That means if $w$ is stabilized by $h$, then $h$ stabilizes the Fourier transform across all irreducible representations. The reverse also holds. 
\end{proposition}

\begin{figure}[htb!]
	\centering
	\includegraphics[width=0.5\textwidth]{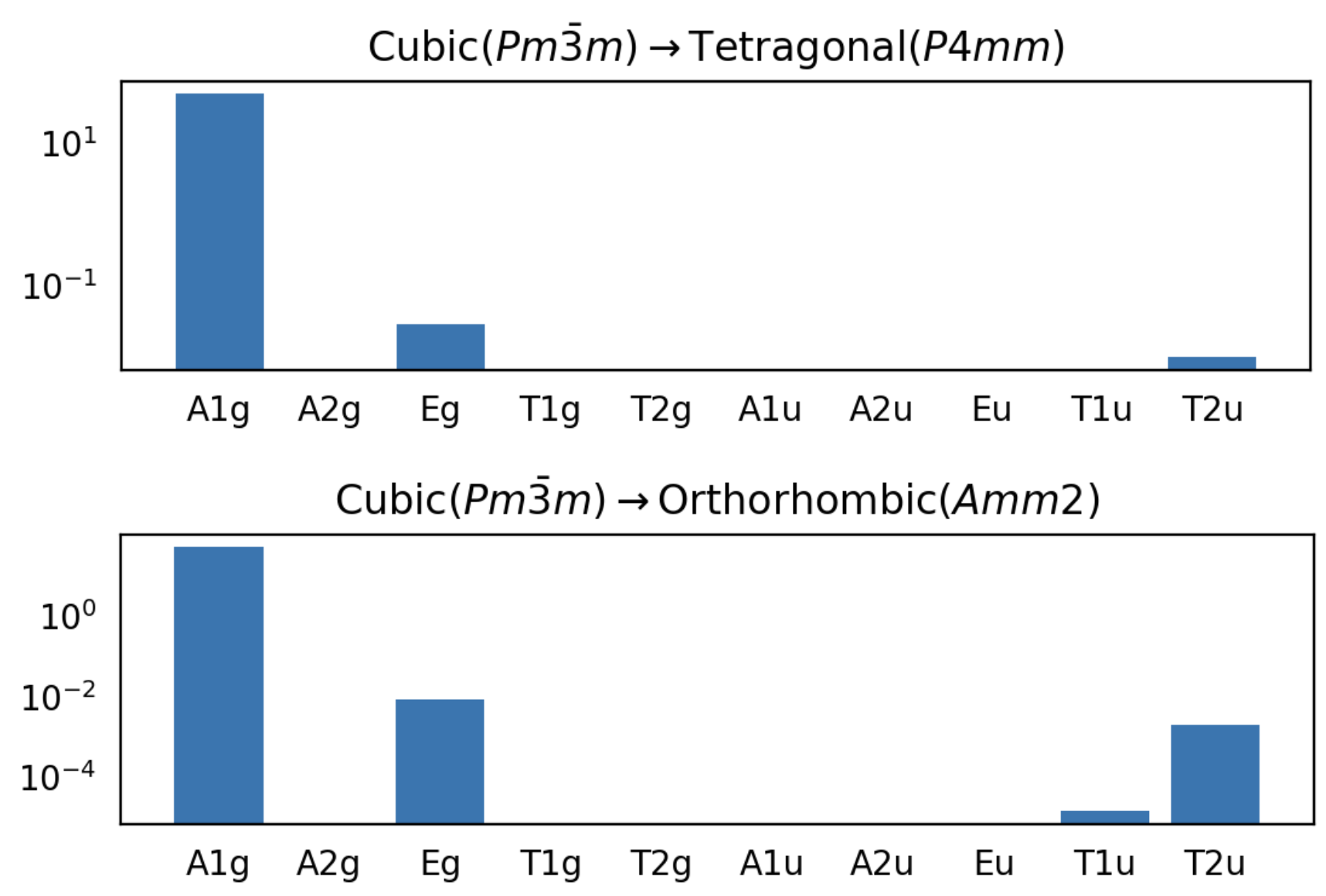}
	\caption{Visualization of the coefficients of 10 irreps of the octahedral group. The symmetries that are preserved after phase transitions are the intersection of the stabilizers of all non-zero irreps.}
	\label{fig:batio3_irrep_coef}
\end{figure}

\section{Turbulence kinetic energy spectrum} \label{app:energy_spectrum}
\begin{figure}[htb!]
	\centering
\includegraphics[width= 0.5\linewidth]{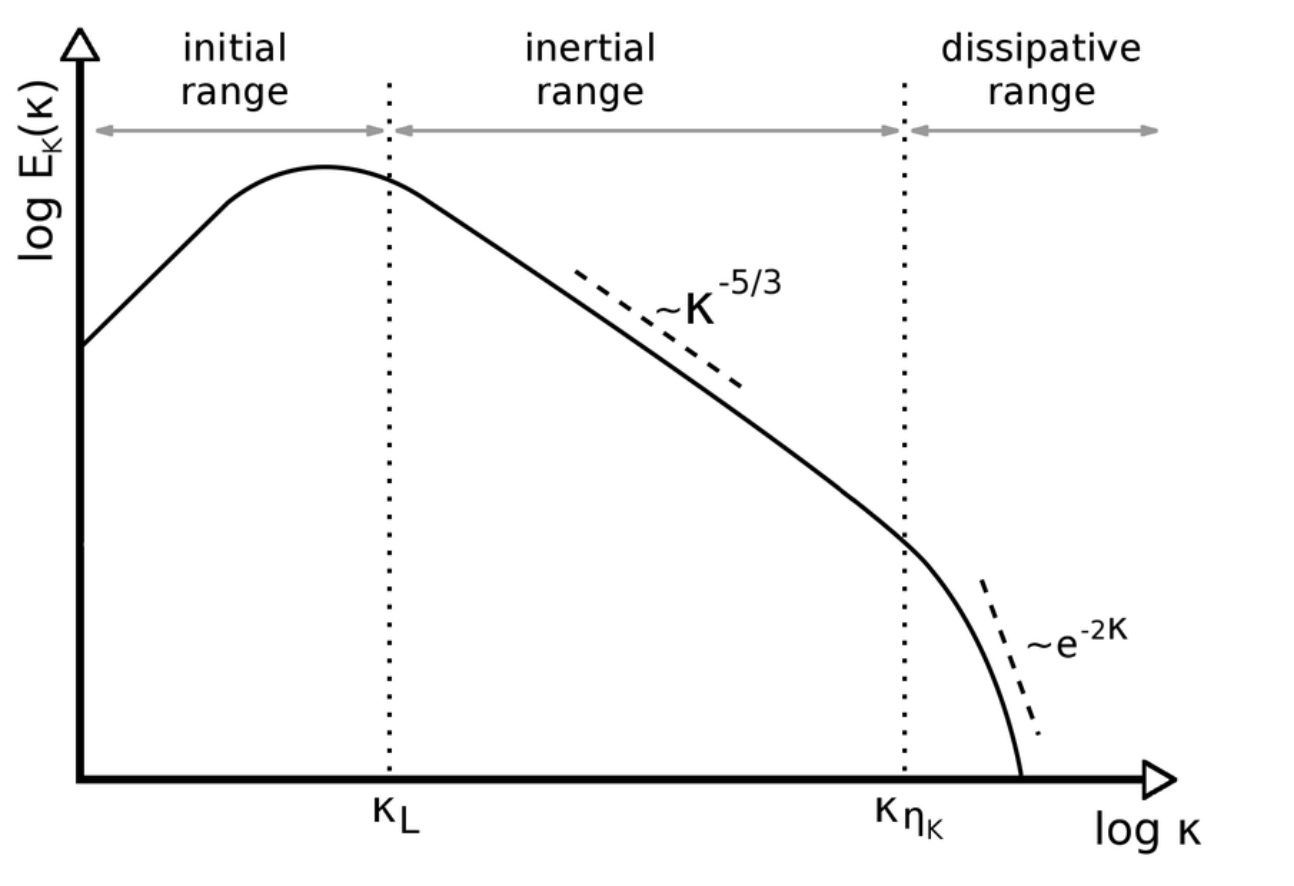}
\caption{Theoretical turbulence energy spectrum plot: The energy spectrum is often represented graphically, showing energy (usually on a logarithmic scale) as a function of wave number or size of the eddies. In Kolmogorov's theory, for the inertial subrange (a range of scales where the energy transfer is dominated by inertia forces rather than viscous forces), the energy spectrum follows a -5/3 power law.}
\label{fig:theoretical_spectrum}
\end{figure}

The turbulence energy spectrum is a concept in fluid dynamics that describes the distribution of kinetic energy among different scales of motion in a turbulent flow. It's based on the understanding that turbulence comprises a wide range of eddies or vortices of different sizes, each carrying a certain amount of energy. The turbulence kinetic energy spectrum $E(k)$ is related to the mean turbulence kinetic energy as 
\begin{align*}
\int_{0}^{\infty}E(k)dk &= (\overline{(u^{'})^2} + \overline{(v^{'})^2})/2, \\ \overline{(u^{'})^2} &= \frac{1}{T}\sum_{t=0}^T(u(t) - \bar{u})^2,
\end{align*}
where the $k$ is the wavenumber and $t$ is the time step. Figure \ref{fig:theoretical_spectrum} shows a theoretical turbulence kinetic energy spectrum plot. The spectrum can describe the transfer of energy from large scales of motion to the small scales and provides a representation of the dependence of energy on frequency. 

\section{Comparison to baselines} \label{app:baselines}
Comparing our method with other symmetry discovery approaches mentioned in section \ref{related_work_baselines} is non-trivial, and these methods fall short of the capabilities demonstrated by our approach for several technical reasons: 1) all of these methods focus on symmetry discovery in the data distribution but lack ability to identify symmetries at the individual sample level; 2) They may only perform well on the datasets with perfect symmetry and lack the quantitative mechanisms to measure the extent of symmetry breaking. 3) They are based on intricate architectures and training strategies and often require careful tuning. 

The works most closely related to our study are MSR \cite{zhou2021metalearning} and LieGAN \cite{yang2023generative}. MSR proposed a method to learn the weight-sharing matrix and the non-constrained filters separately by an optimization-based meta-learning method \cite{finn2017model}. The post-training weight-sharing matrix uncovers the symmetries in the data. We attempted applying MSR to identify rotation and translation symmetries in fluid dynamics. Meanwhile, LieGAN, built on generative adversarial networks, aims to identify the Lie generator of continuous groups from the training data, making it potentially suitable for detecting rotational symmetry in fluid's small eddies

\begin{figure*}[htb!]
	\centering
	\includegraphics[width=0.45\textwidth]{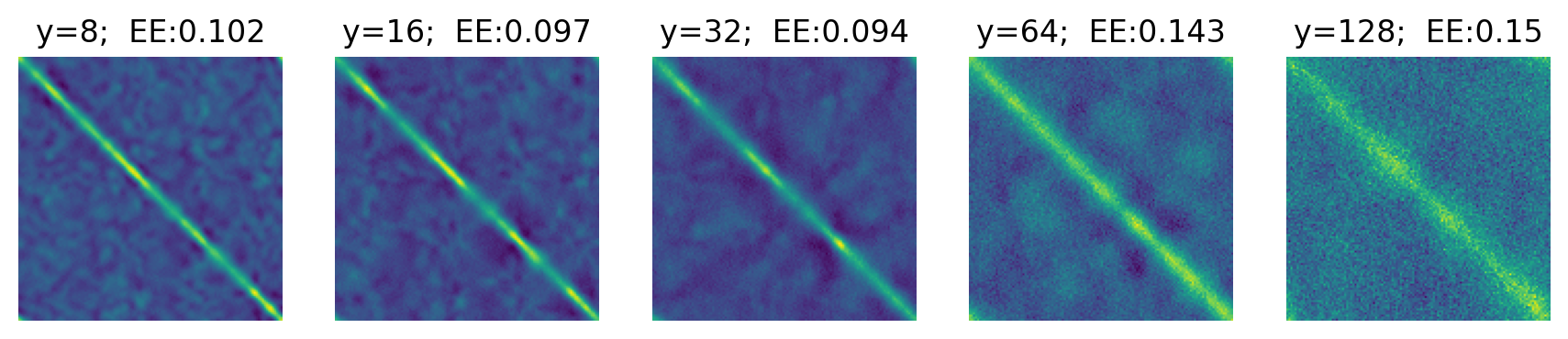}
         \includegraphics[width=0.2\textwidth]{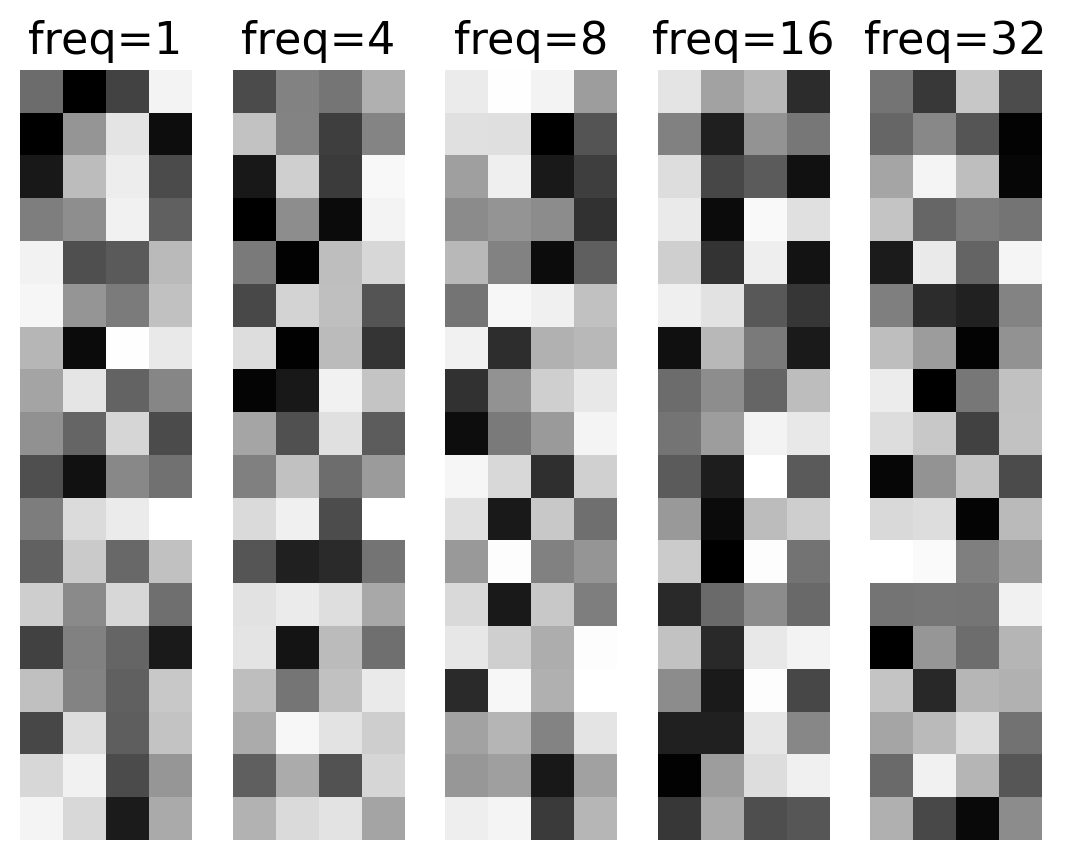}
         \includegraphics[width=0.32\textwidth]{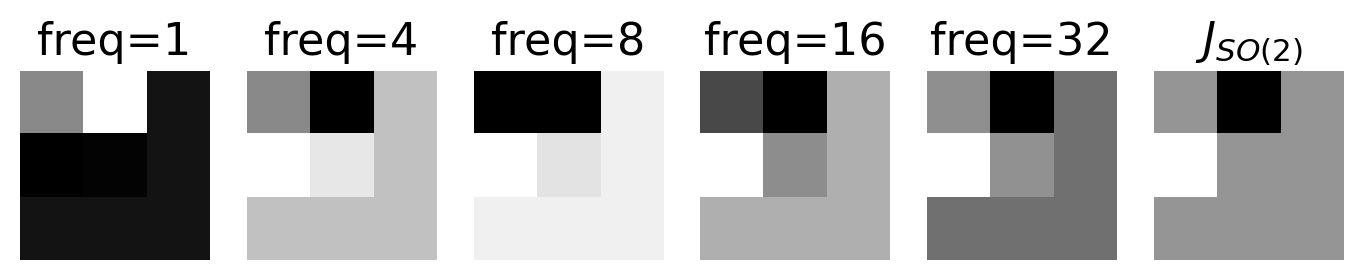}
	\caption{Left: the translation weight-sharing matrices and equivariance errors learned at different y-coordinates through MSR. Middle: learned rotation weight-sharing matrices by MSR at different frequencies of the channel flow. Right: learned SO(2) generators by LieGAN at different frequencies of the channel flow}
	\label{fig:vis_msr}
\end{figure*}

The MSR approach in \cite{zhou2021metalearning} is limited to 1D translation convolution models, which cannot be directly applied to our 2D fluid data. Thus, we train a distinct MSR model for each height (i.e. y coordinate). Figure \ref{fig:vis_msr} left visualizes the translation weight-sharing matrices learned at different y-coordinates through MSR. As we can see, though the weight-sharing matrices are very noisy, the MSR can find that there is a certain amount of translation symmetry in the data. However, when we measured the equivariance errors (EE) of models at different $y$, the results did not accurately reflect the correct amount of symmetries at different heights as we expected. Additionally, we performed a grid search of hyperparameters but failed to make the MSR to learn the correct rotation symmetry from the fluid data.  Figure \ref{fig:vis_msr} middle displays the rotation weight-sharing matrices learned by MSR for the C4 group, where ideally, at high frequencies, the matrix should be a stack of permutation matrices.

Additionally, we conducted experiments with five distinct LieGAN models to capture the rotational symmetry in fluid dynamics across various scales. Initially, using the random initialization prescribed in the original study, LieGAN failed to identify any meaningful generators. Thus, we instead initialized the trainable generator as the true generator of SO(2) group and measured the symmetry breaking by checking how it deviates from the initialization after training. Figure \ref{fig:vis_msr} right visualizes the learned SO(2) generators by LieGAN at five different frequencies of the channel flow. The rightmost one is the true generator of SO2. It seems the generators learned on higher frequencies are closer to it than the ones learned on lower frequencies. However, a critical limitation of LieGAN is its inability to precisely measure the extent of symmetry breaking, unlike our proposed method.


\section{Distributional symmetry breaking with simple forward passing} \label{app:fwdpass}
Assume a probability distribution $p_X: X \to \mathbb{R}$ and a group $G$ that acts on $X$ by $\rho_X$. We have distributional symmetry breaking if $\forall x \in X$, $p_X(x) \neq p_X(\rho_X(g)x)$ (i.e the probability distribution does not remain closed under the group action). In this paper, we present the use of relaxed group convolution to uncover distributional symmetry breaking (i.e. for discovering homogeneity breaking in turbulence). However, we also present a complementary approach showing that group convolutions can detect distributional symmetry breaking on a single forward pass without training equivariant weights. 

In a sequence of group convolutions, one usually averages over the group elements to return a function on $\mathbb{Z}^3$ (or the space of the desired input/output) after the last convolution. To instead detect symmetry breaking, one would not average over the group elements but would rather observe the pattern of the group elements after a single forward pass. Thus, given the output of a group convolution equivariant to a group $H$ as $f_c \in \mathbb{R}^{N \times c _{\text{out}} \times |H| \times \text{spatial dims}}$, we average over the spatial dimensions instead of $|H|$ after the last convolution. The network output will then be in $\mathbb{R}^{N \times c_{out} \times |H|}$. 

\begin{figure}[htb!]
	\centering
	\includegraphics[width=0.7\textwidth]{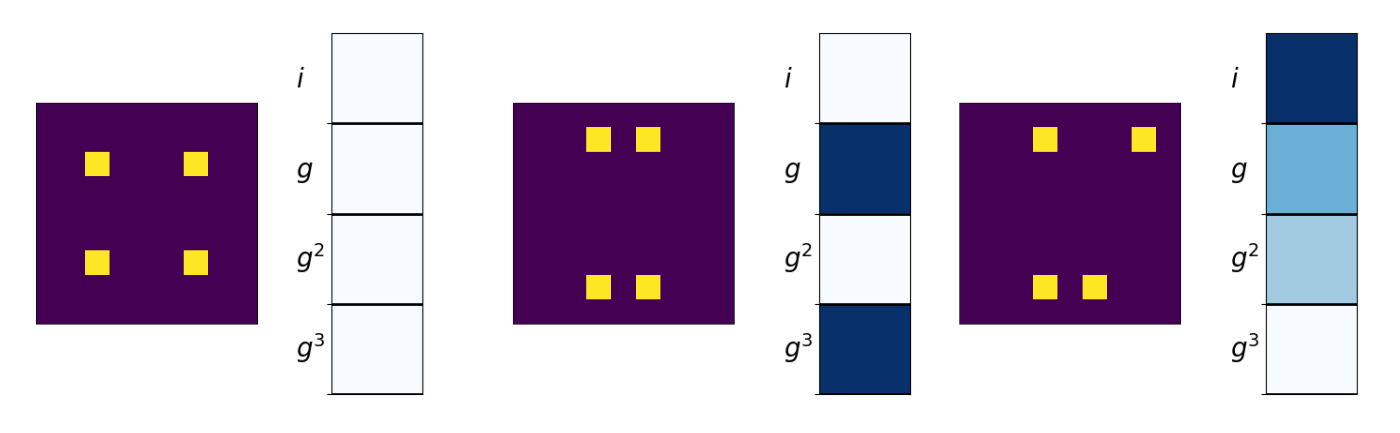}
	\caption{Visualization of input and corresponding output representatation after a single forward pass from a 3-layer $C_4$ equivariant group convolution network for 1) a square 2) a rectangle and 3) a non-symmetric object. These results are shown for a single random model initialization. When observing different model initializations, we found the same patterns for 1) and 2) and the lack of a pattern for 3).}
	\label{fig:fwd_simple}
\end{figure}

We first illustrate this concept with a simple example. We consider a 3-layer $C_4$ equivariant randomly initialized group convolution network and show that the representation after a single forward pass (with no training) reflects the symmetry of the input seen in Figure \ref{fig:fwd_simple}. We observe that when the input is a square with $C_4$ symmetry the output averaged over spatial dimensions remain constant (up to any equivariance error of the model). For a rectangle with $C_2$ symmetry, the elements corresponding to $i$ and $g^2$ are the same while those corresponding to $g$ and $g^3$ are the same, thus also reflecting $C_2$ symmetry. For a shape with no symmetry, the output does not reflect symmetry across group elements.

\begin{figure}
  \centering
  \begin{minipage}[t]{0.48\textwidth}
    \centering
    \includegraphics[width=\linewidth]{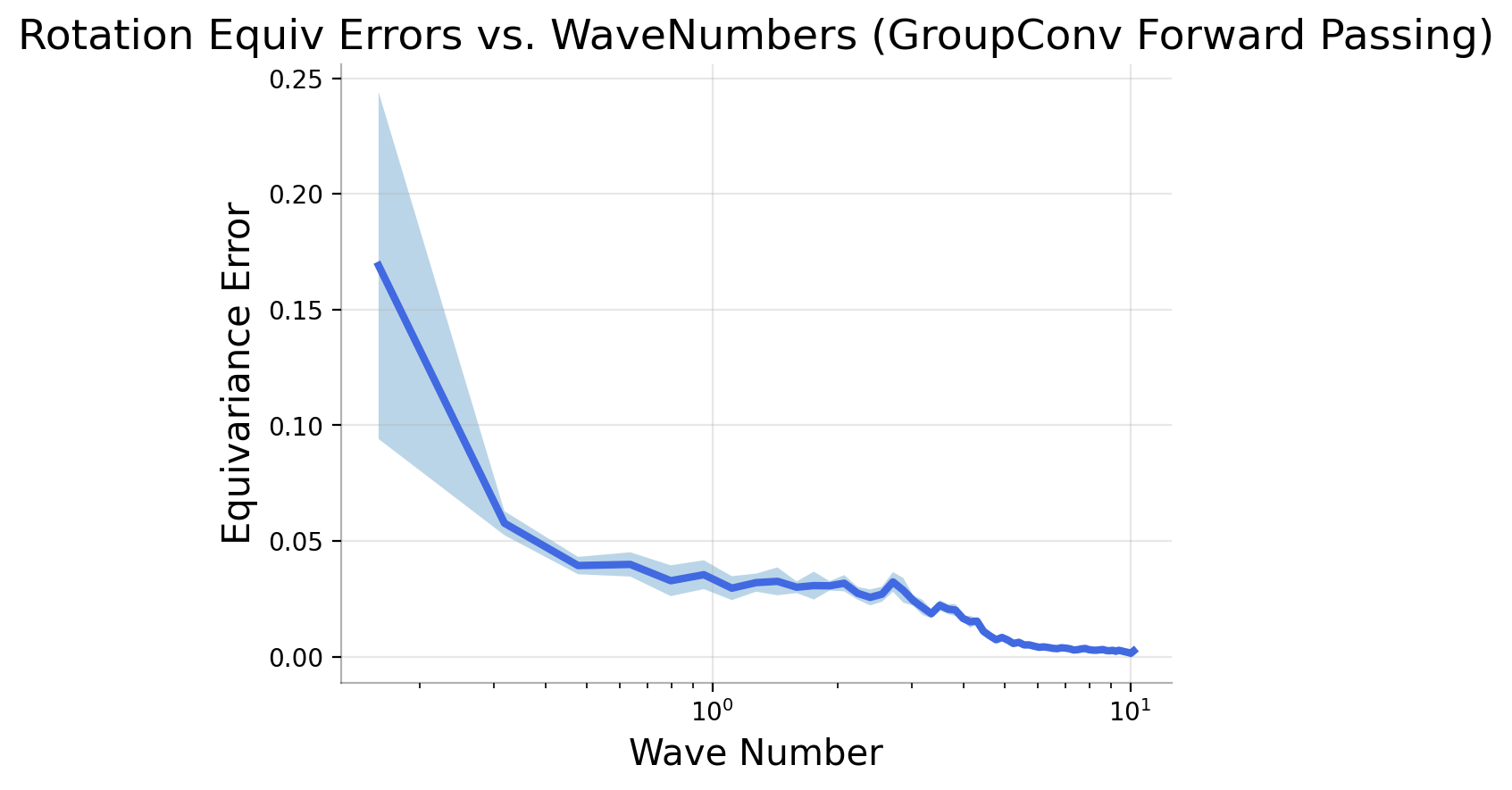}
  \end{minipage}\hfill
  \begin{minipage}[t]{0.48\textwidth}
    \centering
    \includegraphics[width=\linewidth]{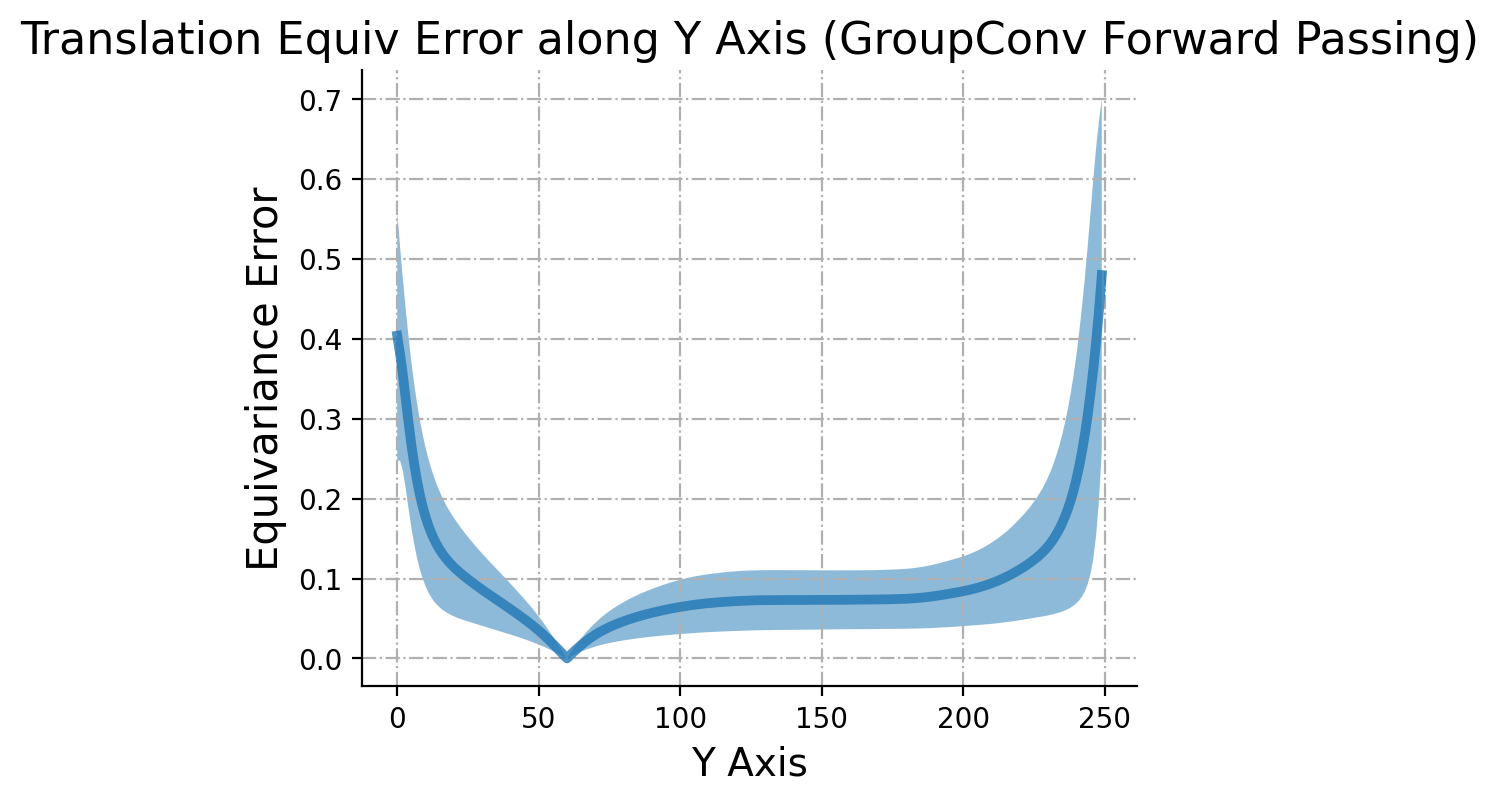}
  \end{minipage}
  \caption{(Left) Rotational equivariance errors vs wavenumber and (right) translational equivariance errors along the y-axis for a single forward pass. The confidence intervals are calculated by 10 randomly initialized models.}
  \label{fig:fwd_turbulence}
\end{figure}

We also explored this method for discovering isotropy breaking and homogeneity breaking in turbulence (Sections \ref{exp:isotropy} and \ref{exp:homogeneity}), as these are more complex examples of distributional symmetry breaking. As shown in \ref{fig:fwd_turbulence}, similar patterns to Figures \ref{fig:isotropy} and \ref{fig:translation} are recovered (up to rescaling of equivariance error). This demonstrates that this method could also be used for symmetry detection in the data distribution.

We emphasize that this method leverages the equivariant representations used in group convolutions to reflect the symmetry of a given input or data distribution. It is thus suitable when one wishes to detect broken symmetries, but could not be used to learn a mapping between input and output (e.g. in the case of functional symmetry breaking). For learning complex dynamics, relaxed group convolution is preferable, as it can automatically identify symmetry breaking and maintain the correct amount of equivariance, enabling end-to-end training with the right inductive bias.


\section{Additional Experiment with Relaxed Weights} 
\begin{figure}[htb!]
	\centering
	\includegraphics[width=0.7\textwidth]{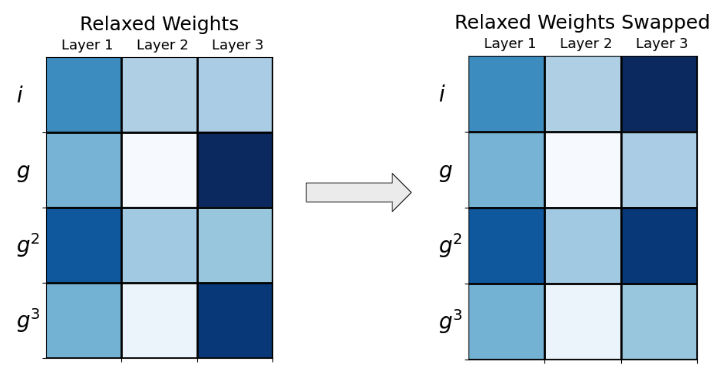}
	\caption{Learned relaxed weights for the $C_4$ model (right) and visualization of swapping the relaxed weights in the last layer.}
	\label{fig:relaxed_weights_buoyant_forces}
\end{figure}

\begin{figure}[htb!]
	\centering
	\includegraphics[width=0.8\textwidth]{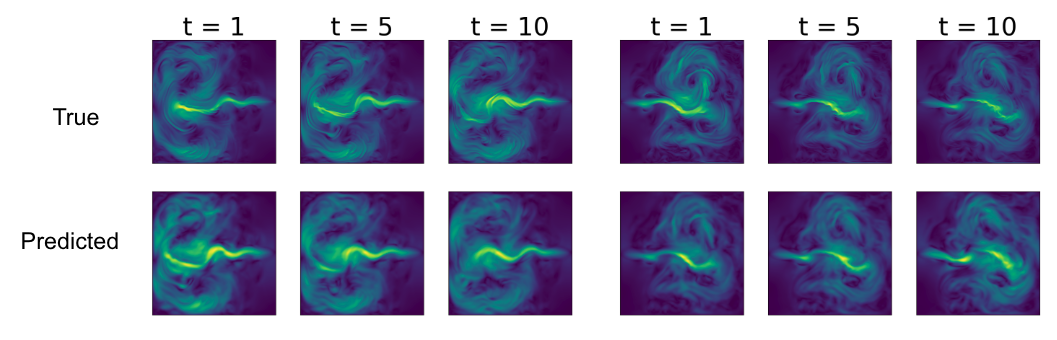}
	\caption{Velocity magnitude for target (ground truth) and model predictions at $t=1,5,10$ in the rotated testing set for simulations in the $C_2$ symmetric dataset. At earlier times $t < 30$, the model performs somewhat comparably to the original model (test RMSE 0.79). However, at later times $30 < t < 50$, the performance degrades (test RMSE = 1.12) while still capturing notable features of the dataset (i.e. the main smoke plume).}
	\label{fig:C2_buoyant_forces_2}
\end{figure}

To further explore the interpretability of relaxed weights, we performed an additional experiment using a 2-D smoke dataset generated with PhiFlow (\citet{phiflow}). In PhiFlow, one can generate simulations with differential initial conditions and external forces from the incompressible Navier Stokes equations. We use a $C_4$ relaxed group convolution model and train it for forward prediction on a dataset with $C_2$ symmetry (2 simulations, one with a buoyancy force in the $+y$ direction and the other with a buoyancy force in the $-y$ direction). Each simulation has 300 timesteps. Steps 0-150 are used for training, 150-200 for validation, and 200-250 for testing. Forecasts are made in an autoregressive manner (using 6 input steps and 4 output steps) and are then evaluated on 10-step ahead prediction RMSEs, as in \citet{wang2022approximately}. Note the model must be trained on these simulations together instead of separately to preserve $C_2$ symmetry. 
\begin{figure}[htb!]
	\centering
	\includegraphics[width=0.8\textwidth]{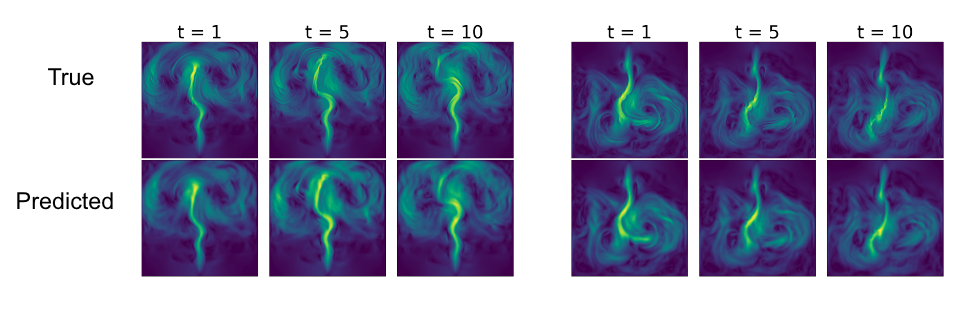}
	\caption{Velocity magnitude for target (ground truth) and model predictions at $t=1,5,10$ in the testing set for simulations in the $C_2$ symmetric dataset. The model seems to accurately capture the dynamics (test RMSE 0.645).}
	\label{fig:C2_buoyant_forces}
\end{figure}
We then consider the learned relaxed weights for the model. As seen in Figure \ref{fig:relaxed_weights_buoyant_forces}, they do exhibit $C_2$ symmetry, as the weights corresponding to the elements $i$ and $g^2$ are the same, as are the weights for the elements $g$ and $g^3$. We consider switching the relaxed weights with the corresponding $C_2$ group elements (i.e. swapping the $g$ and $g^3$ weights with the $i$ and $g^2$ weights). This is shown in Layer 3 in Figure \ref{fig:relaxed_weights_buoyant_forces}.
As the original dataset contains buoyancy forces pointing up/down, we hypothesize that by swapping the relaxed weights of the trained model, we will be able to perform forward prediction on buoyancy forces pointing left/right with the trained model. To do so, we supply the model with an initial start from the testing set shown in Figure \ref{fig:C2_buoyant_forces_2} rotated by 90 degrees. Forecasts are then made autoregressively for 50 timesteps.
By permuting the relaxed weights, we present a ``proof of concept'' forward prediction on rotated simulations reasonably well, demonstrating the interpretability of the relaxed weights. These predictions could potentially be improved by tuning model hyperparameters further or increasing the number of layers/relaxed group convolutions to more accurately capture the dynamics. In the future, we plan to further explore perturbing the relaxed weights for turbulence simulations.

\section{Group and Representation Theory Background} \label{app:background}

We provide a brief informal background on group and representation theory necessary to understand equivariant neural networks and relaxed group convolution. This is quite an extensive subject, so we encourage curious readers to consult \citet{Dresselhaus2008} or other group or representation theory textbooks. We refer the reader to \citet{Cohen2016Group} for the seminal paper introducing group convolutions. Note this background closely follows the notation and definitions presented in \cite{weiler2019e2cnn}.

\textbf{Groups} A group is a mathematical structure composed of a set $G$ and some group action $\cdot, G \times G \to G$. The group must obey the following rules. (i) Closure. $G$ is closed under $\cdot$, $\forall g_1, g_2 \in G, g_1 \cdot g_2 \in G$. (ii) Identity. $\exists e \in G$ such that $eg = ge = g \forall g \in G$. (iii) Inverse. $\forall g \in G, \exists g^{-1} \in G$ such that $g \cdot g^{-1} = e$. (iv) Associativity. $(g_1 \cdot g_2) \cdot g_3 = g_1 \cdot (g_2 \cdot g_3) \forall g_1, g_2, g_3 \in G$. 

\textbf{Representations} More concretely, it is useful to represent abstract group elements by representing them as linear transformations or matrices on some vector space. A representation $\rho$ of a group $G$ on the vector space $\mathbb{R}^n$ is a mapping from $G$ to the general linear group GL($\mathbb{R}^n$) or the group of invertible $n \times n$ matrices that also follows the group structure (i.e. a group homomorphism). 
\begin{align*}
    \rho: G \to \text{GL}(\mathbb{R}^n), \rho(g_1g_2) = \rho(g_1)\rho(g_2) \forall g_1, g_2 \in G, \rho(e) = I
\end{align*}
There can be multiple representations for a single group element. Representations can be combined by taking their direct sum of the corresponding matrices. Concretely, given representations $\rho_1: G \to \text{GL}(\mathbb{R}^n)$ and $\rho_2: G \to \text{GL}(\mathbb{R}^m$, the direct sum $\rho_1 \oplus \rho_2: G \to \text{GL}(\mathbb{R}^{n+m})$ is
\begin{align*}
    (\rho_1 + \rho_2)(g) = \begin{pmatrix}
        \rho_1(g) & 0 \\
        0 & \rho_2(g)
    \end{pmatrix}
\end{align*}

\textbf{Regular Representation} The regular representation is commonly used in equivariant deep learning. The regular representation of a finite group $G$ acts on a vector space $R^{|G|}$. If we associate each basis vector $e_g \in \mathbb{R}^G$ to an element $g \in G$, the representation fo an element $\tilde{g} \in G$ is a permutation matrix that maps $e_g$ to $e_{\tilde{g} \cdot g}$. 

\textbf{Irreducible Representation} Irreducible representations are representations that do not contain a smaller representation. Formally, consider a representation $\rho : G \to \text{GL}(V)$ of a group $G$ where $V$ is some vector space. A linear subspace $W \in V$ is $G$-invariant if $\rho(g)w \in W, \forall g \in G, \forall w \in W$. If a representation contains a $G$-invariant subspace, then there exists a change of basis to an equivalent representation which can be decomposed into the direct sum of independent representations on the invariant subspace and its orthogonal complement. A representation is irreducible if no non-trivial invariant subspace exists. A representation of a compact group $G$ can thus be decomposed as 
\begin{align*}
    \rho(g) = Q \left [\bigoplus_{i \in I} \psi_i(g) \right ] Q^{-1}
\end{align*}
For finite groups, an irreducible representation cannot be decomposed in this block diagonal form.

\textbf{Group and Representation Theory in Materials} We present some background that is relevant for the experiments discovering symmetry breaking factors in phase transitions of $BaTiO_3$ in Section \ref{exp:crystal}. Given a group $G$, two elements $g$ and $g'$ are conjugate if there exists another element $x \in G$ such that $g' = x^{-1}g x$. A class is the set of group elements that can be obtained from a given element $g \in G$ by conjugation. The elements of a given group can thus be divided into classes. Classes correspond to physically distinct kinds of symmetry \cite{Dresselhaus2008} and are thus useful for categorizing different symmetries of a material rather than enumerating all individual group elements as in Figure \ref{fig:batio3_all_weights}. See \citep{Dresselhaus2008, zee2016group} for more information on group/representation theory in physics and materials.


\end{document}